%% file: main.tex
\documentclass[10pt,twocolumn,letterpaper]{article}

\usepackage{cvpr}              %

\input{preamble}

\definecolor{cvprblue}{rgb}{0.21,0.49,0.74}
\usepackage[pagebackref,breaklinks,colorlinks,citecolor=cvprblue]{hyperref}

\def\paperID{147} %
\def\confName{3DV\xspace}
\def\confYear{2026\xspace}

\title{M2SVid: End-to-End Inpainting and Refinement for Monocular-to-Stereo \\Video Conversion}

\author{
Nina Shvetsova$^{1,2,3,4}$\thanks{This work was conducted during an internship at Google.}, Goutam Bhat$^{1}$, Prune Truong$^{1}$, Hilde Kuehne$^{2,3}$,  Federico Tombari$^{1,5}$\\
\small $^{1}$Google,
\small $^{2}$Tuebingen AI Center/University of Tuebingen, 
\small $^{3}$Goethe University Frankfurt, \\
\small $^{4}$MPI for Informatics, Saarland Informatics Campus, 
\small $^{5}$Technical University of Munich \\ 
\vspace{-3mm} \\
Project webpage: \url{https://m2svid.github.io/}
}

\begin{document}
\maketitle
\input{sec/0_abstract}    
\input{sec/1_intro}

\input{sec/2_related_work}

\input{sec/3_method}

\input{sec/4_experiments}

\input{sec/5_conclusion}

{
    \small
    \bibliographystyle{ieeenat_fullname}
    \bibliography{main}
}

\clearpage

\addcontentsline{toc}{section}{Appendix} 
\appendix
\noindent{\Large\bf Supplementary Material}\\[1em]

\input{sec/supplementary}

\end{document}

%% file: preamble.tex
\usepackage{graphicx}
\usepackage{caption}
\usepackage{subcaption}
\usepackage{amsmath}
\usepackage{amsfonts}
\usepackage{booktabs}
\usepackage{siunitx}
\usepackage{wrapfig}
\usepackage{bbm}
\usepackage{multirow}
\usepackage{relsize}
\usepackage{color}
\usepackage{colortbl}
\usepackage{algorithm}
\usepackage{listings}

\usepackage{enumitem}
\usepackage{url}
\usepackage{xspace}
\definecolor{almond}{rgb}{0.94, 0.87, 0.8}

\newcommand{\parsection}[1]{\vspace{1mm}\noindent\textbf{#1:}~}
\newcommand{\parpoint}[1]{\vspace{1mm}\noindent\textbf{#1}~}
\newcommand{\ours}{M2SVid\xspace}
\newcommand{\vect}[1]{\boldsymbol{\mathbf{#1}}}

\definecolor{BrightOrange}{RGB}{255,165,0}
\newcommand{\fix}[1]{{#1}}

\usepackage{pifont}
\newcommand{\cmark}{\textcolor{green}{\ding{51}}} %
\newcommand{\xmark}{\textcolor{red}{\ding{55}}}   %

%% file: sec/0_abstract.tex
\begin{abstract}

We tackle the problem of monocular-to-stereo video conversion and propose a novel architecture for inpainting and refinement of the warped right view obtained by depth-based reprojection of the input left view. 
We extend the Stable Video Diffusion (SVD) model to utilize the input left video, the warped right video, and the disocclusion masks as conditioning input to generate a high-quality right camera view. In order to effectively exploit information from neighboring frames for inpainting, we modify the attention layers in SVD to compute full attention for discoccluded pixels. Our model is trained to generate the right view video in an end-to-end manner without iterative diffusion steps by minimizing image space losses to ensure high-quality generation. 
Our approach outperforms previous state-of-the-art methods, being ranked best $2.6\times$ more often than the second-place method in a user study, while being $6\times$ faster.\footnote{To be published at 3DV 2026. 
When citing this work, please refer to the final version published in IEEE Xplore. Cite as:
Nina Shvetsova, Goutam Bhat, Prune Truong, Hilde Kuehne, Federico Tombari.
``M2SVid: End-to-End Inpainting and Refinement for Monocular-to-Stereo Video Conversion''.
In \textit{International Conference on 3D Vision}, IEEE, 2026.
}

\end{abstract}

%% file: sec/1_intro.tex
\section{Introduction}
\label{sec:intro}

Emerging technologies such as Mixed-Reality headsets and glasses allows users to easily experience immersive content, thanks to the use of separate displays for left and right eyes.
Rendering videos from left and right eye viewpoints on these displays creates a stereoscopic 3D effect, giving viewers a sense of depth.
However, capturing such stereoscopic videos usually requires specialized cameras that can capture both left and right eye perspectives simultaneously.
While manual monocular-to-stereo video conversion is possible, it is costly and time-consuming. This drives the need for automated, fast, and high-quality conversion methods  that can enable large scale conversion of videos.

\input{figures/teaser}

Recent advancements in video generation~\cite{chen2024videocrafter2,blattmann2023stable} as well as monocular depth estimation~\cite{depth_anything_v2,ke2024repurposing} have led to significant improvements in the monocular-to-stereo conversion quality.
A common approach is to use depth maps generated by a monocular depth model to warp the input video to a virtual right camera and then inpaint the disoccluded regions~\cite{mehl2024stereo,dai2024svg,shi2024stereocrafter}. SVG~\cite{dai2024svg} repurposes a pre-trained video diffusion model for the inpainting task, while ~\cite{mehl2024stereo} trains a custom model.
Rather than only modifying discoccluded regions, the recent concurrent work StereoCrafter~\cite{shi2024stereocrafter} finetunes a video diffusion model~\cite{Blattmann2023StableVD} to inpaint and refine the full warped video, using only the warped view and discocclusion masks as input. 
While obtaining improved results, StereoCrafter still leverages an architecture primarily designed for monocular video inpainting task, without exploiting the constraints specific to stereoscopic video-refinement.
Moreover, the inference requires many diffusion steps 
leading to high computational cost.

In this work, we propose a novel architecture for efficient inpainting and refinement in the monocular-to-stereo conversion task. 
Our approach is designed to leverage important additional cues available in this task. Firstly, in the areas that are not disoccluded, the right view largely resembles the left one, where the image content is just horizontally shifted.  Thus, even if depth-based warping introduces artifacts in the generated right view, we conjecture that most of them can be fixed by relying on left view information.
Secondly, the disoccluded regions are generally small in most cases. Furthermore in the presence of camera or scene motion, the disoccluded regions in one frame are often visible at nearby frames, thus simplifying the inpainting task into a spatial-temporal matching problem. The aforementioned aspects motivate us to design \ours  (\textbf{M}onocular-\textbf{to}-\textbf{S}tereo \textbf{Vid}eo conversion), a specialized architecture which performs inpainting and refinement in an feed-forward manner, \textit{without any iterative diffusion steps.}

Our main contributions are the following. First, we extend the Stable Video Diffusion (SVD) architecture by employing \textit{the input left view}, in addition to the warped right view and disocclusion mask as conditioning for the refinement task. This makes it easier for the model to propagate information from the uncorrupted left view to the target right view, thus preserving high frequency details such as fine structures. Secondly, rather than performing temporal attention only over the same spatial locations as in SVD~\cite{Blattmann2023StableVD}, we compute \textit{full cross-attention for disoccluded pixels.} This provides greater flexibility to propagate useful information from either nearby temporal frames or the associated left view, with a limited computational overhead. Thirdly, we train our model in an  \textit{end-to-end manner} using perceptual and fidelity based losses, enabling high quality inpainting and refinement (See Fig.~\ref{fig:teaser}). %
Finally, unlike other methods, we train on publicly available datasets, making our approach reproducible and enabling fairer comparisons.

We perform quantitative evaluation, qualitative analysis, as well as a human perception user studies to evaluate our method. It shows that our \ours outperforms previous state-of-the-art StereoCrafter and SVG, being ranked best $2.6\times$ more often than StereoCrafter and $4.75\times$ more often than SVG, while running $6\times$ and $600\times$ faster, respectively.

%% file: figures/teaser.tex
\begin{figure}[t]
\centering
\vspace{-4mm}
\includegraphics[width=1\linewidth]{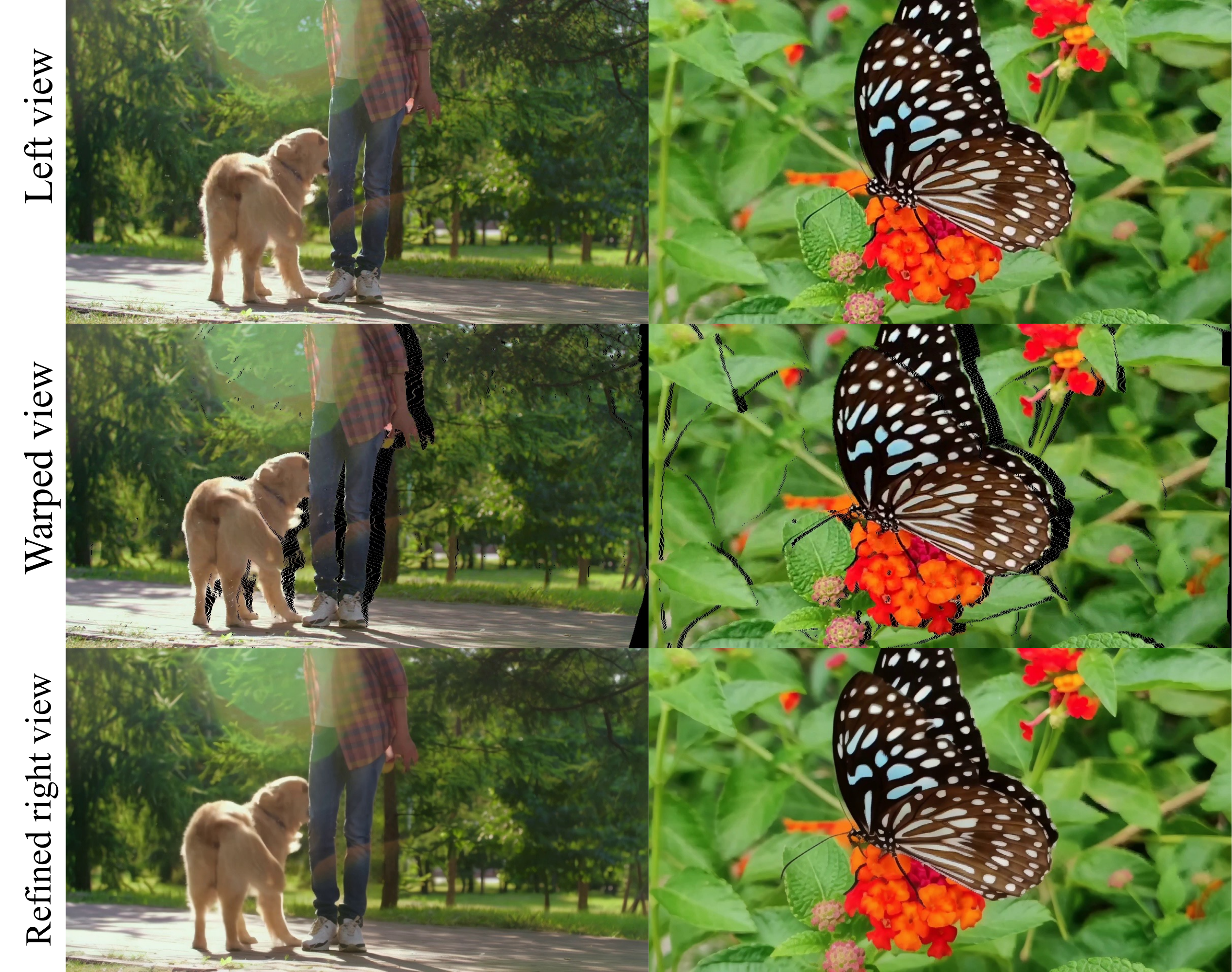}
\vspace{-6mm}
\caption{\small{
We present M2SVid, an end-to-end video inpainting and refinement approach for monocular-to-stereo video conversion task. Given an initial right view, obtained via \eg depth based warping, our method inpaints the missing region and refines the artifacts introduced by warping in a feed-forward manner. 
\label{fig:teaser}
}}
\vspace{-6mm}
\end{figure}

%% file: sec/2_related_work.tex
\section{Related Work}
\label{sec:related_work}

\subsection{Novel View Synthesis}
Early methods for novel-view synthetis performed 3D reconstruction of the full scene given dense input views~\cite{colmap,KarHM17,YaoLLFQ18}, which can then be used to render novel views. While Multi-Plane Image (MPI)-based approaches~\cite{tucker2020single,han2022single,single_view_mpi} were successfully used to generate multiplane representations from single images, NeRFs~\cite{mildenhall2020nerf} marked a milestone in the field, and spawned many efforts to improve quality~\cite{mipnerf,mipnerf360}, reduce latency~\cite{mueller2022instant,HedmanSMBD21,Nex,PlenOctrees,NeffSPKMCKS21}, extend to large-scale scenes~\cite{Turki_2022_CVPR}, explicit representation~\cite{kerbl20233d}, pose-free setup~\cite{barf, nerfmm, GARF, GNerF, sinerf, sparf}, or fewer inputs~\cite{dietnerf, infonerf, Regnerf, DSNerf, densedepth,sparf, reconfusion}, including just a single image~\cite{Melas-KyriaziL023,LiuXJCTXS23,LiuSCZXW00G024,LiuWHTZV23}.
Approaches such as CAT3D~\cite{gao2024cat3d}, NVS-Diffusion~\cite{Karras2024edm2}, MultiDiff~\cite{Muller_2024_CVPR} directly generate the queried camera view using a diffusion models.
A few works also extend these to videos~\cite{gao2024cat4d}. While obtaining promising results, the generated views can still have artifacts due to the inherent difficulties of rendering views with large viewpoint changes. %

\subsection{Monocular-to-Stereo Conversion}
Unlike general novel-view synthesis, monocular-to-stereo conversion renders a fixed right view from a left view, enabling monocular content to be experienced in 3D on mixed-reality headsets.
Most prior works can be grouped into two families. One line of work aims to directly generate the right image/video given the left input~\cite{xie2016deep3d,xie2016deep3d,shi2024immersepro,Lee2017Automatic2C}. A notable example is Deep3D~\cite{xie2016deep3d}, which leverages an internal disparity representation to directly render the right video given the left one.
The second type of approaches employ an explicit depth map to reproject the left image to the right camera, and then inpaint the disoccluded regions~\cite{Shih3DP20,Jampani2021SLIDESI,mehl2024stereo,dai2024svg,shi2024stereocrafter,konrad2013learning,wang2023learning,dai2024svg}.
Notably, Wang~\etal~\cite{wang2023learning} learn a diffusion model for inpainting disoccluded pixels using self-supervision with random cycle rendering. SVG~\cite{dai2024svg} uses a pre-trained video diffusion model for inpainting. While the method can achieve spatio-temporal consistency in the output, the inpainting result can be incorrect due to the lack of task-specific fine-tuning. Mehl~\etal~\cite{mehl2024stereo} aim to instead perform a geometry aware inpainting of the disoccluded areas using local background information, rather than generating arbitrarily realistic content. 
StereoCrafter~\cite{shi2024stereocrafter}, which is concurrent to our work, fine-tunes a video diffusion model for inpainting the disoccluded areas, on a large stereo dataset (not released). %
Another recent work SpatialDreamer~\cite{spatialdreamer} mitigates the necessity of paired stereoscopic training data by proposing a self-supervised training framework using a forward-backward rendering  mechanism. 
Note that a number of the above methods do not release code or model, and are often trained on private collected data, making a fair comparison difficult.

\subsection{Diffusion Models}
Diffusion Models~\cite{ho2020denoising,song2020denoising,rombach2022high} 
are generative models that iteratively denoise an input to produce the output.
They have been tremendously successful for the text-to-image~\cite{ho2020denoising,rombach2022high,Saharia2022PhotorealisticTD,Zhang2023AddingCC,Ramesh2022HierarchicalTI} and text-to-video~\cite{blattmann2023stable,Ho2022ImagenVH} tasks. They have also been successfully employed for diverse computer vision tasks, thanks to the strong image priors learned by the models. These include
image~\cite{Lugmayr2022RePaintIU,Corneanu2024LatentPaintII,Liu2023ImageIV} and video~\cite{zhang2023avid,cherel2023infusion,lee2025videoinpainter} inpainting, novel view synthesis~\cite{Muller_2024_CVPR,kalischek2025cubediffrepurposingdiffusionbasedimage}, as well as 
monocular depth estimation~\cite{ke2024repurposing,hu2024depthcrafter,garcia2024fine,shao2024learningtemporallyconsistentvideo}, semantic segmentation~\cite{Kawano2024MaskDiffusionEP}, or normal estimation~\cite{fu2024geowizard}. Conventionally, diffusion models employ multiple denoising steps during inference, leading to large computations cost. A number of approaches aim to train one-step models to address this, using knowledge distillation~\cite{song2024sdxs,Zhang2023HiPAEO}, GAN training~\cite{Mao2024OSVOS}, low-rank adaptors~\cite{Zhang2023HiPAEO}. Garcia~\etal~\cite{garcia2024fine} instead finetune a pre-trained diffusion model to perform direct feed-forward monocular depth estimation using end-to-end training. In this work, we show that such an end-to-end training strategy can also be employed for inpainting disoccluded regions in monocular-to-stereo video conversion task.

%% file: sec/3_method.tex
\section{Conditional Latent Diffusion Models}
\label{sec:diffusions}

Here, we provide a brief background on diffusion models employed in our method. Denoising Diffusion Probabilistic Models (DDPMs)~\cite{ho2020denoising} are generative models trained to map a simple noise distribution $p_T$ to the data distribution $p_0$, by reversing a stochastic forward process $p_t$, $t = 1, ..., T$. A denoising model $\hat{\vect{v}}_{\theta}(x_t, t)$ is trained to generate an image from pure noise by progressively denoising inputs $x_t$ at time stamp $t$. In order to reduce computational complexity, Latent Diffusion Models (LDMs)~\cite{rombach2022high} operate in a latent space of a Variational Autoencoder (VAE)~\cite{kingma2013auto}.
Conditional diffusion models~\cite{saharia2022photorealistic, zhang2023adding} extend the denoising model $\hat{\vect{v}}_{\theta}(x_t, t, c)$ to be conditioned on additional input $c$, such as text~\cite{saharia2022photorealistic}, images~\cite{saharia2022photorealistic}, \etc to control the generation process.

\parsection{Training and inference with conditional LDMs} 
During training, given a data sample $ \mathbf{x} $ (e.g., an image or video) and its corresponding conditioning input $c$ (e.g., text or another image), the data sample $ \mathbf{x} $ is first encoded into a latent representation $ \mathbf{z} = E(\mathbf{x}) $ using VAE encoder $ E $. 
The latent representation $\mathbf{z}$ is perturbed through a forward diffusion process:
$\mathbf{z}_t = \sqrt{\bar\alpha_t} \mathbf{z} + \sqrt{1 - \bar\alpha_t} \boldsymbol{\epsilon}$
where $\boldsymbol{\epsilon} \sim \mathcal{N}(0, I)$ is Gaussian noise, and $\bar\alpha_t$ controls the noise schedule. Instead of predicting the noise $\boldsymbol{\epsilon}$ directly, diffusion models usually leverage \textit{v-parameterization}~\cite{salimans2022progressive}, where the model $\hat{\vect{v}}_{\theta}(\mathbf{z}_t, t, c)$ learns to predict:
$\vect{v} = \alpha_t \boldsymbol{\epsilon} - \sqrt{1 - \bar\alpha_t}  \mathbf{z}.$
Therefore the model $\hat{\vect{v}}_{\theta}$, typically a U-Net~\cite{ronneberger2015u}, is trained to reconstruct $\vect{v}$, by optimizing the objective:
\begin{equation}
\label{eq:diffusion_loss}
    \mathcal{L} = \mathbb{E}_{\mathbf{z}, c, \boldsymbol{\epsilon} \sim \mathcal{N}(0,I), t \sim \mathcal{U}(T)} \left[ \| \vect{v} - \hat{\vect{v}}_{\theta}(\mathbf{z}_t, t, c) \|^2 \right].
\end{equation}

During inference, the denoising process starts from pure noise $\vect{z}_T$ and the learned denoiser $\vect{v}_{\theta}(z_t, t, c)$ iteratively refines the output over $T$ steps to recover the final sample.

\parsection{Diffusion as feed-forward models} 
Garcia~\etal~\cite{garcia2024fine} propose to use pre-trained diffusion U-Net model as a feed-forward models for pixel-to-pixel tasks such as depth and normal estimation. In this scenario, the timestep $t$ is not sampled anymore and fixed as $t=T$ to train the model for single-step prediction. The noise $\boldsymbol{\epsilon}$ is additionally replaced by the mean of the noise distribution, \ie, zero. With $t=T$, we get $\bar\alpha_T\approx0$, and thus $
\mathbf{z}_T = \sqrt{\bar\alpha_T} \mathbf{z} + \sqrt{1 - \bar\alpha_T} \boldsymbol{\epsilon} \approx \mathbf{0}$ and $\vect{v} = \alpha_t \boldsymbol{\epsilon} - \sqrt{1 - \bar\alpha_t}  \mathbf{z} \approx - \mathbf{z}.$
As such, the model directly learns to predict the clean output $\vect{v} \approx - \mathbf{z}$ starting from zero noise vector $\mathbf{z}_T = \mathbf{0}$. With such timestep sampling and assumptions, the standard loss (\cref{eq:diffusion_loss}) converges to the $L_2$ loss in the latent space between the ground truth $ - \mathbf{z}$ and the predicted latents $\hat{\vect{v}}_{\theta}(\mathbf{0}, T, c)$:
\begin{equation}\label{eq:latent}
    \mathcal{L}_{latent} = \mathbb{E}_{\mathbf{z}, c} \left[ \| ( (-\mathbf{z}) - \hat{\vect{v}}_{\theta}(\mathbf{0}, T, c) \|^2 \right].
\end{equation}

However, \cite{garcia2024fine} proposes decoding the predicted latents, $\mathbf{\hat{z}} = -\hat{\vect{v}}_{\theta}(\mathbf{0}, T, c)$, using a VAE decoder $D$ to obtain the reconstructed output 
$\mathbf{\hat{x}} = D(\mathbf{\hat{z}})$. Instead of minimizing the latent space error~\cref{eq:latent}, they propose to optimize a task-specific loss directly in image space with $\mathcal{L} = L_{task}(\mathbf{x}, \mathbf{\hat{x}})$.

\begin{figure}
\centering

\includegraphics[width=0.99\columnwidth]{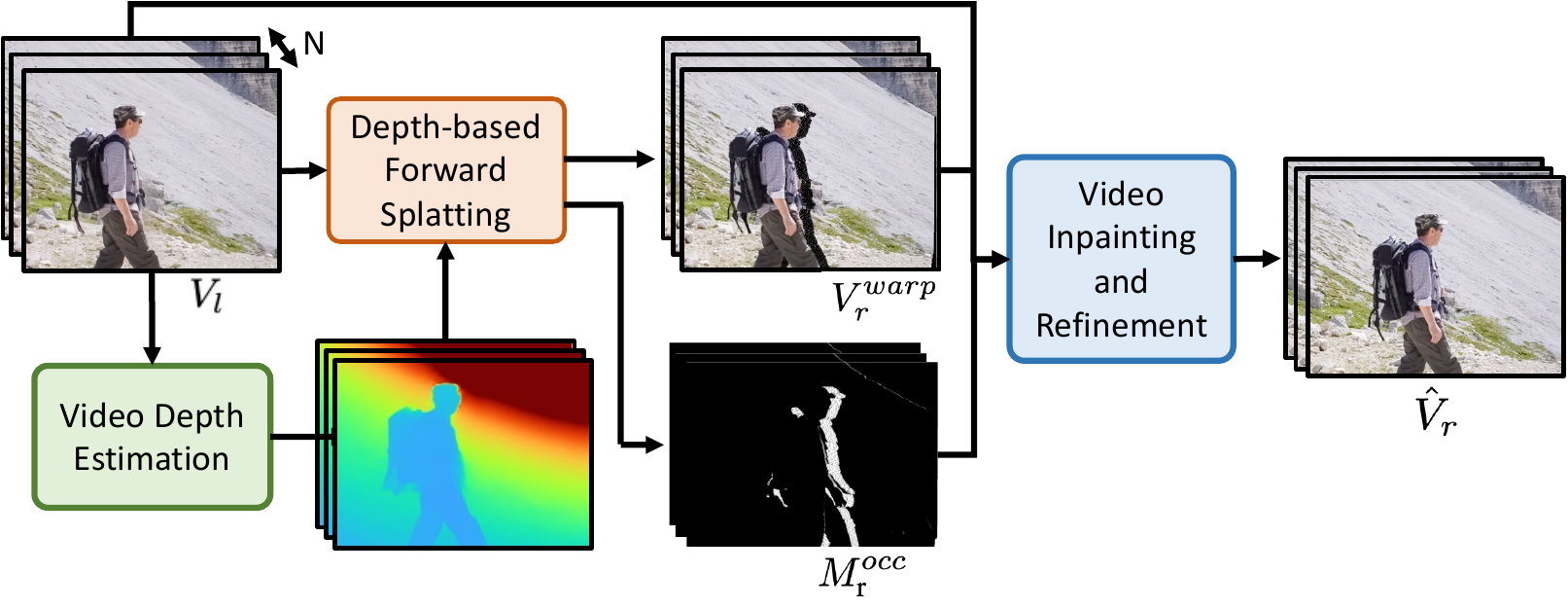}
\vspace{-4mm}
\caption{\small{Overview of our monocular-to-stereo conversion pipeline. Given an input monocular video, we first estimate per-pixel depth, which is used to warp the input video to a right camera view. The input video, the warped video, as well as the disocclusion masks are then passed to our video inpainting and refinement module to generate the final right view.
        \textbf{
        \label{tab:overview}
        }}}
\vspace{-5mm}
\end{figure}

\section{Monocular-to-Stereo Conversion Pipeline}
\label{sec:pipeline}
In this work, we tackle the problem of converting any 2D video to a 3D stereoscopic video. Given a monocular video $V \in \mathbb{R}^{N \times H \times W \times 3}$, containing N frames of resolution $H \times W$, the goal is to transform it into a \textit{stereoscopic video pair} $(V_{l}, V_{r})$, where $V_{l}, V_{r} \in \mathbb{R}^{N \times H \times W \times 3}$ represent the views corresponding to the left and right eyes, respectively. Following prior work, we assume that the input video $V$ corresponds to the left-eye view, i.e., $V_{l} = V$, and therefore, the goal is to only generate the right-eye view $V_{r}$ from $V$. Note that multiple possible right-eye views $V_{r}$ exist depending on the baseline, \ie the distance between cameras. Hence we condition the generation on a maximum desired disparity $D_\text{max}$ between $V_{l}$ and $V_{r}$, that 
can be set by the user to control the stereoscopic effect in practical applications.

An overview of the pipeline is shown in Fig.~\ref{tab:overview}. We follow the commonly employed strategy~\cite{mehl2024stereo,shi2024stereocrafter} of performing depth-based reprojection to generate an estimate of the right view, as described next.

\parsection{Video-depth estimation} We use a monocular video depth estimation model to obtain per-frame depths. Our method is agnostic to the choice of depth estimation method. 

\parsection{Left-to-right video warping} We use the user-provided maximum disparity $D_\text{max}$ to convert the depth maps to disparities. The disparity maps are then used to forward-splat the input left-view video to a virtual right view denoted as $V_{r}^{warp}$. Note that the warped right frames inevitably contain missing values (\ie holes) introduced by disocclusions. These missing pixels are indicated by the disocclusion mask $M_{r}^{occ}$, obtained as a by-product of the splatting.

\parsection{Video inpainting and refinement} The final step in our pipeline is to generate a high quality right view given the warped video $V_{r}^{warp}$. As mentioned before, the warped right video $V_{r}^{warp}$ contains holes introduced by disocclusions.
Furthermore, the warped right video $V_{r}^{warp}$ can also contain artifacts introduced due to reprojection and depth errors. 
In this work, we mainly focus on this final ``video inpainting and refinement" task. To this end, we introduce an efficient end-to-end model to inpaint the disoccluded regions and fix reprojection artifacts. Our architecture is described in detail in the next section.

\section{End-to-End Stereoscopic Video Refinement}\label{method}

We introduce \ours, an efficient end-to-end model for generating a high-quality \textit{right video} from the \textit{warped right video} $V_{r}^{warp}$. While this problem is similar to the standard video inpainting task, it comes with a set of unique properties, as described next. 

\parpoint{\textbf{(1)}} 
In our setup for rendering a right camera view, inpainting holes appear in regions where depth increases from left to right. These regions correspond to background, so unlike general-purpose inpainting (e.g., object removal), our model can rely on context from only one side of the hole.

\parpoint{\textbf{(2)}} In a general inpainting setup, the model is only required to generate the missing pixels, while keeping the rest of the image unaltered. 
In our case, however, regions outside the inpainting mask may also contain artifacts from depth-map errors or interpolation during forward splatting.
These artifacts are especially common in presence of thin structures such as fences, where the estimated depth can have substantial errors. Fortunately, we can leverage the original left video to correct the errors introduced by the reprojection.

\parpoint{\textbf{(3)}} 
In the image inpainting task, the model \textit{must hallucinate} content in missing regions due to the lack of additional information.
A video inpainting model can instead use information from multiple frames, though large holes still require substantial hallucination.
In our task, however, the inpainting regions are much thinner
(i.e., as controlled by the maximum disparity $D_\text{max}$). As a result, the inpainting problem is greatly simplified and the model \textit{can copy} information from other temporal frames to avoid hallucination.

\noindent We develop a custom architecture for the right view generation task aiming to exploit the aforementioned properties.

\input{figures/method}

\subsection{Overview}\label{subsec:arch}
An overview of our method is shown in Fig.~\ref{fig:method}. We base our model on a strong pre-trained video diffusion model to benefit from learned video priors. In particular, we utilize Stable Video Diffusion (SVD)~\cite{Blattmann2023StableVD}, a latent video diffusion model  trained to generate videos from an input image. We customize the SVD architecture for the stereoscopic video generation task as follows. First, we condition each latent generation on the input left video $V_{l}$, reprojected right video $V_{r}^{warp}$, as well as the disocclusion masks $M_{r}^{occ}$. 
Secondly, only for the inpainted pixels, we extend the spatial attention in SVD to operate over all pixels in all frames, instead of just the pixels in the same spatial location. This gives greater flexibility for the model to copy visible pixels from neighboring frames in order to consistently inpaint.
Finally, we train our model \textit{end-to-end} on public stereoscopic datasets to generate the refined right view in a \textit{feed-forward manner}, without requiring multiple denoising steps.

\subsection{High-Frequency Details Preservation}\label{left-cond}

Conventional video inpainting methods utilize masked video to condition inpainting models, a strategy widely adopted in stereo conversion methods~\cite{shi2024stereocrafter,mehl2024stereo,wang2023learning}, where the inpainting model is conditioned solely on the warped right video $V_{r}^{warp}$ and the disocclusion mask $M_\text{r}^\text{occ}$.
However, as discussed in~\cref{method} (2), the warped right video is likely to contain artifacts even in non-inpainting regions. Thus we propose to also utilize the original left video $V_\text{l}$ as an extra input to the model. In more detail, the standard SVD model takes the noisy latents for a video snippet and the VAE encoding of a conditioning image as input. %
We modify the SVD architecture to instead take the following inputs as conditioning: the VAE encoding of the left video snippet $E(V_{l})$, the VAE encoding of the warped right snippet $E(V_{r}^{warp})$, and the disocclusion mask $M_\text{r}^{occ,resized}$, resized to the same resolution as the VAE encodings. We thus denote the model $\hat{\vect{v}}_{\theta}(\vect{z}_t, t, c)$ with conditioning inputs 
$c$ as
\begin{equation}\label{eq:conditioning}
 c = [E(V_{l}), E(V_{r}^{warp}), M_\text{r}^{occ,resized}] \,\,,
\end{equation}
where $[.]$ refers to concatenation. This modification is achieved by modifying the first convolution layer in the U-Net, as is the common practise. 

Using the original left video as conditioning allows the model to easily infer high frequency details as well as other information which might have been destroyed during the depth based reprojection stage. As shown in Fig.~\ref{fig:alb_left_cond}, this 
improves the quality of the generated right views, with high-frequency details from the left view better preserved.

\subsection{Spatio-Temporal Aggregation for Inpainting}\label{full-attention}
The SVD model takes multiple video frames (\ie a snippet) as input and jointly denoise them to produce a temporally consistent output. Ideally, one would compute a full attention over all spatial tokens in all frames for maximal information flow. 
However, this would be prohibitively expensive, resulting in a complexity of
$H=N^2 \times h^2\times w^2$ 
for each attention layer, where $h$ and $w$ are the size of the latent representations.
For this reason, the standard video diffusion models such as SVD and VideoCrafter~\cite{chen2024videocrafter2} factorize the full attention computation into interleaved spatial and temporal attention layers.
In the spatial attention layers, only tokens from the same timestamp attend to each other, while in the temporal attention layers, tokens from the same spatial location attend to each other across different timestamps. This drastically reduces the complexity to 
$N \times h^2 \times w^2 + N^2 \times h \times w$, 
which is more palatable.

However, factorizing full attention can reduce modeling capacity for our task, especially in dynamic scenes with camera motion. As discussed in \cref{method}(3), an occluded pixel in one frame is often visible at a different spatial location in another, simplifying inpainting problem. In such cases, we would prefer full spatio-temporal attention.
Fortunately, the number of inpainted tokens constitute only a small fraction of all tokens (usually $<$ 5\%). 
We exploit this property and modify the spatial attention layer in SVD to allow tokens corresponding to the disocclusion mask $M^{occ}_{r}$ to attend to all other tokens, while using the spatial attention only for other tokens. This improves the inpainting capability of our model (Fig.~\ref{fig:alb_full_attention}) without significant computational overhead.

\input{figures/qualitative}

\subsection{Efficient Feed-Forward Prediction}\label{sec:single-step-training}
The diffusion models for text-to-image/video generation and inpainting commonly utilize multiple denoising steps to generate the final image~\cite{ho2020denoising,saharia2022photorealistic,rombach2022high}. This is important due to the ill-posed nature of the problem wherein multiple solutions exists for a single input prompt. Utilizing fewer denoising steps usually results in blurry output. 
However, in our task of right view generation, the inpainting regions are generally small. Furthermore, as described in Sec.~\ref{method}, the neighboring temporal frames as well as the input left frames contain a significant amount of information needed to generate a high-quality, temporally consistent right video. The model therefore does not need to perform significant hallucination in most cases, but rather needs to fetch the relevant information from other frames. The relatively constrained nature of our problem thus motivates us to generate the right view in a direct feed-forward manner, rather than the multi-step denoising used in prior work~\cite{dai2024svg,shi2024stereocrafter}, similar to the approach employed by Garcia~\etal~\cite{garcia2024fine} for depth estimation.

Given the input left video $V_\text{l}$, warped right video $V_\text{r}^{warp}$, and the disocclusion masks $M_\text{r}^{occ}$, we generate the conditioning input $c$ to the model by obtaining the VAE encodings of left and reprojected videos, as shown in~\cref{eq:conditioning}. The initial latent is set to the mean noise \textbf{0} as in~\cite{garcia2024fine} \fix{and the timestamp is set to the highest
noise $t=T$. As discussed in~\cref{sec:diffusions}, this choice of initial latent and timestamp leads the Video U-Net to directly reconstruct a clean latent due to the use of \textit{v-parameterization}~\cite{salimans2022progressive}.} Therefore, the conditioning input, together with the initial latent are then passed through the Video U-Net $\hat{\vect{v}}_{\theta}$ to obtain the latent encoding $\hat{z} = - \hat{\vect{v}}_{\theta}(\textbf{0}, T, c)$ of the right video. This is then passed through the VAE decoder $D$ to obtain the predicted right view $\hat{V}_{r}$. The inference pipeline can thus be denoted as, 
\begin{equation}
\label{eq:single_pass_pred}
    \hat{V}_{r} = D(- \hat{\vect{v}}_{\theta}(\textbf{0}, T, c))
\end{equation}
Our feed-forward prediction strategy significantly reduces the latency of the right view generation step compared to existing methods, as shown in Table~\ref{tab:desktop_study}. Furthermore, this also allows us to train the model end-to-end using image quality losses \wrt to the ground truth right video $V_{r}$, in order to maximize the quality of the generated video.
In particular, we train the model using a combination of LPIPS and $L_\text{1}$ losses directly in the image space, along with an auxiliary loss~\cref{eq:latent} in the latent space. Our final training loss is thus defined as,
\begin{equation}
\label{eq:training_loss}
    \mathcal{L} = \mathcal{L}_{latent}(z, \hat{z}) + \mathcal{L}_{\text{L1}}(V_{r}, \hat{V}_{r}) + \mathcal{L}_{\text{LPIPS}}(V_{r}, \hat{V}_{r})
\end{equation}
Here $z$ corresponds to the VAE encodings of the ground truth right video $V_{r}$. Note that during training, we keep the VAE decoder frozen and only fine-tune the U-Net model.

\subsection{Training Data}
\label{sec:data_preprocessing}
To train our model using the supervised objective~\cref{eq:training_loss}, we require a dataset stereoscopic videos with their corresponding disparity maps.
Due to the lack of large-scale stereoscopic datasets, prior works focused on collecting private datasets that usually have not been publicly released. Instead we aim to train using publicly available datasets.

We utilize \textbf{Ego4D}~\cite{grauman2022ego4d} and \textbf{Stereo4D}~\cite{jin2024stereo4d} datasets for our training. Ego4D a large-scale egocentric video dataset containing approximately 263 long videos (80 hours in total) collected using a stereo camera, while the recently released \textbf{Stereo4D}~\cite{jin2024stereo4d} dataset consists of $\sim$ 200K stereoscopic video clips sourced from $\sim$ 7K online videos. The Stereo4D dataset also provides rectified videos and disparity maps.
The Ego4D dataset on the other hand contains unrectified videos, without any precomputed disparity maps.

We thus perform the following steps to preprocess Ego4D dataset.
First, we uniformly sample frames and perform dense feature matching with LoFTR~\cite{sun2021loftr}, using these matches to compute the fundamental matrix~\cite{Zhang2021} via RANSAC~\cite{Fishchler}. Next, we estimate rectification transformations for both views and rectify the videos. We then compute the LoFTR matches again and shift the left and right views horizontally until all disparities between the matched points are positive and the smallest disparity is zero.
This ensures that the 
videos follow a rectified stereo setup.
Finally, we use an off-the-shelf stereo matching method BiDaVideo \cite{jing2024match-stereo-videos} to obtain disparity maps for all stereo pairs.

%% file: figures/method.tex
\begin{figure*}
\centering
\includegraphics[width=0.90\linewidth]{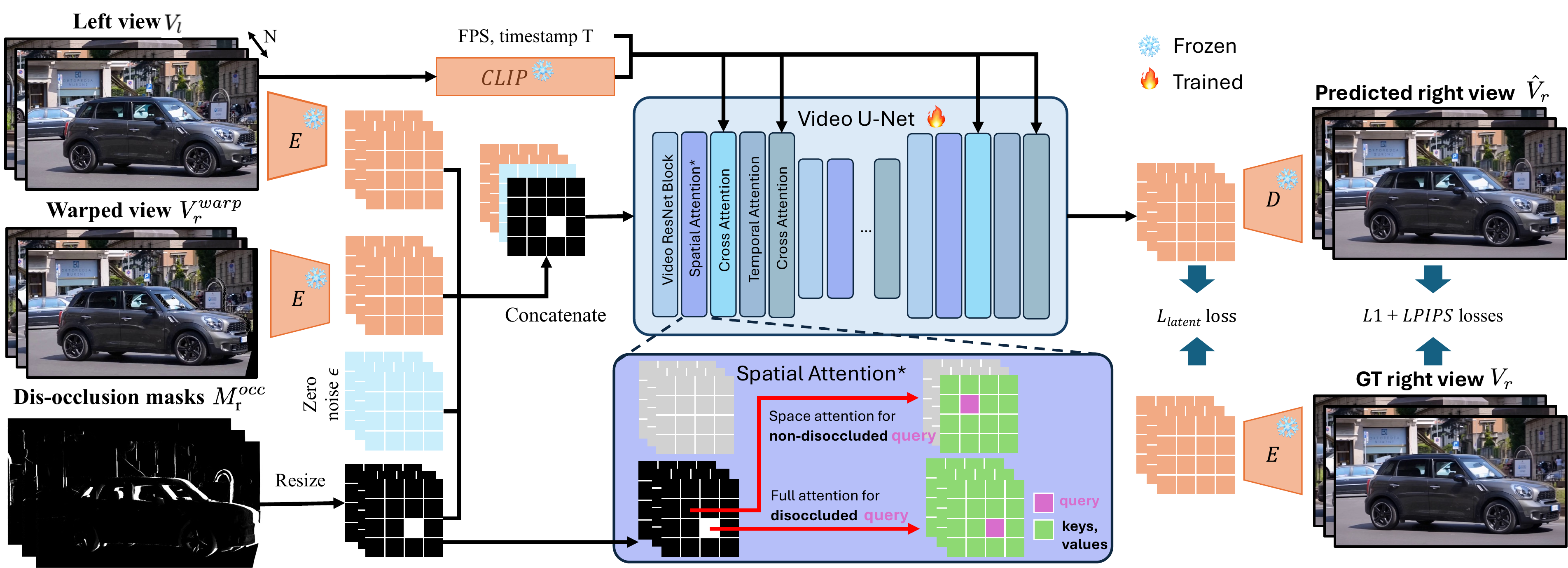}
\caption{\small{An overview of our proposed stereoscopic video refinement method. Our model inpaints the discoccluded regions in the warped right view, and corrects possible artifacts introduced by warping errors. The model takes the VAE encodings of the input left view, reprojected right view, and the disocclusion mask as conditioning to the U-Net. The latent encodings of the refined right view are then generated in a single denoising step, and then decoded by the VAE Decoder to generate the output right video. In order to effectively utilize the information from neighboring frames for inpainting, we extend the spatial attention layer in SVD to compute full attention for the disoccluded tokens. The model is training end-to-end by minimizing image space and latent space losses.
        \textbf{
        \label{fig:method}
        }}}
\vspace{-4mm}
\end{figure*}

%% file: figures/qualitative.tex
\begin{figure*}
\centering
\includegraphics[width=0.85\linewidth]{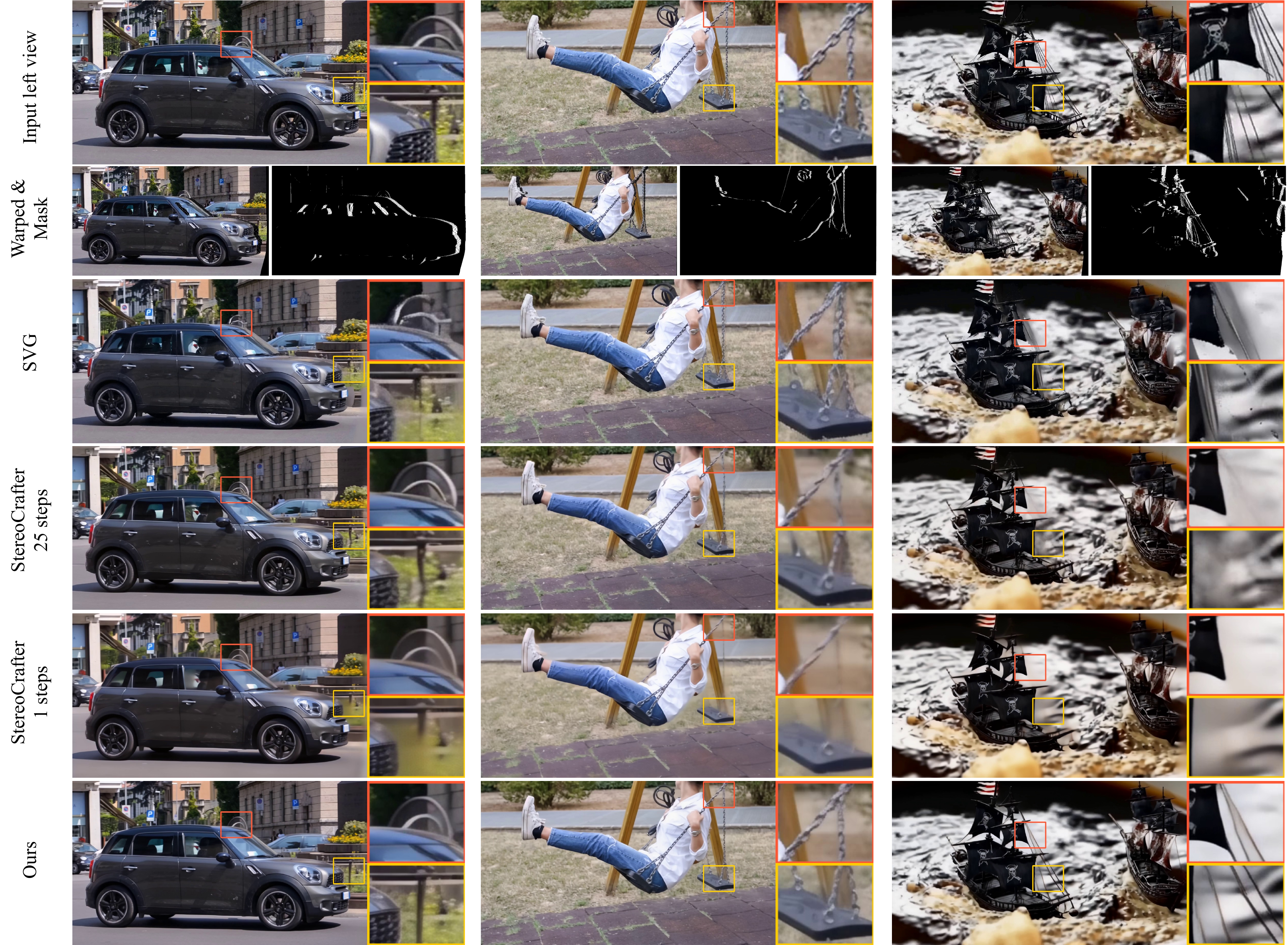}
\vspace{-2mm}
\caption{\small{Qualitative comparison of our approach with state-of-the-art methods SVG~\cite{dai2024svg} and StereoCrafter~\cite{shi2024stereocrafter}. Our approach can effectively preserve the high-frequency information from the input video and generate high-quality right views.
        \textbf{
        \label{fig:qualitative_sota}
        }}}
\vspace{-3.5mm}
\end{figure*}

%% file: sec/4_experiments.tex
\section{Experimental Results}
\label{sec:experimental_evaluation}

In this section, we evaluate the quality of the right view videos generated by our method, both qualitatively as well as quantitatively. 
Further results, analysis, visualizations and implementation details are provided in the Appendix.

\parsection{Quantitative evaluation} To quantitatively evaluate our approach, we need datasets containing left and right stereo videos, together with the (pseudo) ground truth disparity maps which are used to generate the warped right views for each method.
This allows us to directly compare the generated right views with the ground truth right views, without having to worry about errors introduced by incorrect depth estimation. We rely on \textbf{Stereo4D}~\cite{jin2024stereo4d} and \textbf{Ego4D}~\cite{grauman2022ego4d}  test sets and 
use the standard image quality metrics PSNR, MS-SSIM, and LPIPS~\cite{zhang2018perceptual} for evaluation. We compute the metrics independently over each video, and average them over the full dataset. 
In the appendix, we also report metrics inside and outside disoccluded regions.
All evaluations are performed on 16-frame videos, sampled at 5 FPS, resized to 512 resolution, and centrally cropped.

\input{tables/sota}

\input{tables/desktop_study}
\input{tables/headset_study}

\parsection{Qualitative evaluation} For qualitative analysis and user studies, we only need monocular videos along with per-frame depth maps. We use videos from the \textbf{DAVIS} dataset~\cite{perazzi2016benchmark} as well as videos from free online stock sources\footnote{\url{https://www.pexels.com/}}. We sample 16 frames per video at 8 FPS and resize them to $640\times1152$ resolution. We compute the depth maps using the recent DepthCrafter~\cite{hu2024depthcrafter} method.

\subsection{Comparison to State-of-the-Art}

\parsection{Baselines} 
We compare our method to the recent state-of-the-art monocular-to-stereo conversion models \textbf{SVG}~\cite{dai2024svg} and \textbf{StereoCrafter}~\cite{shi2024stereocrafter}. 
SVG is a training-free method that utilizes a frozen diffusion model to inpaint regions within the disocclusion mask while not modifying the remaining regions. StereoCrafter, which is an unpublished concurrent work, fine-tunes SVD~1.1~\cite{Blattmann2023StableVD} to inpaint and refine the warped right views in a diffusion-based denoising setup, using the warped right views along with the disocclusion masks as conditioning. We evaluate StereoCrafter using the default 25 denoising steps as well as single denoising step. We exclude the Deep3D~\cite{xie2016deep3d} from the evaluation as it doesn't allow control over the baseline between the stereo cameras and is shown to have inferior results~\cite{shi2024stereocrafter,dai2024svg} compared to SVG and StereoCrafter. 
In order to ensure a fair comparison, we use the same depth maps for reprojection in all methods. 
However, we use the official warping implementation provided by the authors for each method.

\parsection{Qualitative Results} We perform a qualitative comparison with the SVG and StereoCrafter methods in~\cref{fig:qualitative_sota}.
Since SVG only performs inpainting within the disocclusion mask, it fails to fix the artifacts introduced by errors in warping.
Furthermore, it can incorrectly extend the `foreground' object to fill the inpainting hole due to lack of task-specific fine-tuning.
StereoCrafter with a single denoising step produces blurry inpainting and degrades the quality of the warped areas as well. This is expected since the model was trained using a multi-step denoising objective.
When using 25 denoising steps, StereoCrafter inference generally produces sharp results. However it can struggle at times to correctly generate the high-frequency details. This is because the high-frequency information can often get degraded during the warping step. Since StereoCrafter only utilizes warped view and masks as conditioning, it can struggle to recover the details.  
In contrast, our model is conditioned on the input left view as well, allowing it to leverage the full context for inpainting and refinement. 
Furthermore, since our method is trained in an end-to-end manner with image space losses, it can learn to minimize the loss of high-frequency information introduced by VAE decoder.

\parsection{Quantitative Results} 
We quantitatively compare our approach with SVG and StereoCrafter on Stereo4D in~\cref{tab:sota}. 
Despite using only a single-step inference, \ours significantly outperforms both baselines on all metrics, except for SVG on MS-SSIM, where SVG benefits in MS-SSIM from preserving non-empty pixels (warped using ground truth disparity), while our method incurs a slight MS-SSIM drop due to VAE compression, but achieves superior visual quality by correcting warping errors, e.g., in thin structures.

\parsection{User studies} We also perform two user studies.
First, in a \textit{desktop} user study,
participants viewed a random subset of 21 videos from the DAVIS dataset and public sources.
For each video, anonymized right views from each method were shown alongside the input left view and disocclusion mask (as reference). 
Each of the 13 participants were asked to rank the quality of the generated videos from 1 (best) to 4 (worst), taking into account factors such as temporal consistency, image quality (i.e. sharpness) and lack of artifacts. \fix{If methods are indistinguishable, the equal ranking was allowed.} 
In total, 112 rankings per method were collected (\cref{tab:desktop_study}). 
Our method significantly outperforms all others, obtaining an average ranking of 1.43, compared to 2.05 obtained by StereoCrafter (25 denoising steps) and 2.88 by SVG.  \fix{In fact, our method was ranked first $2.6\times$ more often than StereoCrafter (25 steps), and $4.75\times$ than SVG. }

\fix{To validate that the enhanced visual quality of our method also improves user experience in stereoscopic viewing, we conducted a second user study using a \textit{VR headset}. Participants were shown anonymized stereo videos generated by our method and StereoCrafter (25 denoising steps) and asked which version they preferred, or if both were equal. We collected 105 comparisons from 5 users (21 each). As shown in~\cref{tab:headset_study}, our method was preferred in 39 cases, while StereoCrafter was favored in only 9 (the rest are tied), highlighting the clear advantage of our approach.}

\input{figures/ablations}

\parsection{Run-time} 
Furthermore, our efficient refinement achieves a 6$\times$ and 635$\times$ speed-up compared to StereoCrafter with 25 steps and SVG, respectively, on an A100 GPU (Tab.~\ref{tab:desktop_study}).

\subsection{Ablation Study}
We ablate the key components of our approach in this section on the Ego4D and Stereo4D datasets.

\parsection{Left view conditioning (Sec.~\ref{left-cond})} The impact of using the left view as conditioning, in addition to using the warped view and disocclusion mask is shown in~\cref{tab:abl_cond}. Conditioning on the left video shows a 6.0\%, 5.8\% and 9.8\% improvement in PSNR, MS-SSIM and LPIPS metrics respectively on Ego4D with 25 denoising steps. A similar improvement is observed in the feed-forward case. This is because the left video conditioning allows the model to recover high-frequency details and correct artifacts introduced during the warping, as seen in Fig.~\ref{fig:alb_left_cond}.

\input{tables/ablations/conditioning}

\input{tables/ablations/end_to_end}

\parsection{Full-attention at disoccluded pixels (Sec.~\ref{full-attention})} While this contribution doesn't impact metrics in~\cref{tab:abl_cond} (likely
due to the limited number of complex scenes in the test set), our qualitative analysis shows benefits for dynamic scenes with camera motion,
where both foreground and background move. In~\cref{fig:alb_full_attention}, we observe 
that our method prevents hallucinations (top example) and enables correct inpainting (bottom) of thin structures. More results are in Appendix.

\parsection{End-to-end training (Sec.~\ref{sec:single-step-training})} In Tab.~\ref{tab:abl_end_to_end}, we ablate our proposed end-to-end training strategy. When comparing the model trained with the standard diffusion loss (\cref{eq:diffusion_loss}) with 25 (A) or 1 (B) diffusion step at inference, the single step model leads to more blurry results than (A), as evidenced by the 12\% and 20\% decrease in LPIPS on Stereo4D  and Ego4D respectively. End-to-end training with only latent space supervision~(\cref{eq:latent}) (C) only slightly improves the LPIPS metric. Our final training loss (\cref{eq:training_loss}) (D), including LPIPS and L1 losses in image space, largely improves the image sharpness, as evidenced by the 24\% and 22\% improvement in LPIPS on Stereo4D  and Ego4D compared to (C). Notably, we outperform the standard diffusion loss with 25 inference steps (A) on all metrics and datasets.

%% file: tables/sota.tex
\begin{table}[]
    
    \centering
    \resizebox{1\columnwidth}{!}{
    \begin{tabular}{@{}lcc|ccc }
    	\toprule
                Method & Training data & Denoising steps & PSNR $\uparrow$ &  MS-SSIM $\uparrow$ &  LPIPS $\downarrow$  \\
                \midrule
                SVG~\cite{dai2024svg} & - (training free) & 50 steps & 25.6 & \textbf{0.926} & 0.217 \\
                StereoCrafter~\cite{shi2024stereocrafter} & private dataset  & 25 steps &24.9 & 0.909 & 0.242 \\
                StereoCrafter~\cite{shi2024stereocrafter} & private dataset  &1 step & 25.3 & 0.911 & 0.262 \\
               \rowcolor{almond} \ours (Ours) & Stereo4D + Ego4D & 1 step &\textbf{26.2} & 0.915 & \textbf{0.180}  \\
                \bottomrule
    \end{tabular}
     }
      \vspace{-0.3cm}
    \caption{\small{State-of-the-art comparison on Stereo4D test set. Our approach obtains the best scores in terms of PSNR as well as LPIPS.
        \textbf{
        \label{tab:sota}
        }}}
    \vspace{-0.7cm}
\end{table}

%% file: tables/desktop_study.tex
\begin{table}[]
    
    \centering
    \resizebox{1\linewidth}{!}{
    \begin{tabular}{@{}l|c|c|c|c@{}}
    	\toprule
                \multirow{2}{*}{Method}  & Denoising  & \multirow{2}{*}{Average rank$\downarrow$}  & \fix{\# Chosen best}  & \multirow{2}{*}{Latency (s) $\downarrow$} \\
                
                  &  steps & & \fix{(rank=1) $\uparrow$} &  \\
                
                \midrule
                SVG~\cite{dai2024svg} & 50 steps &  2.88 & 16 / 112 & 1270.4  \\
                StereoCrafter~\cite{shi2024stereocrafter} & 25 steps &  \underline{2.05} & \underline{29 / 112} & 12.2 \\
                StereoCrafter~\cite{shi2024stereocrafter} & 1 step & 3.46 & 4 / 112 & \underline{2.4}  \\
               \rowcolor{almond} \ours (Ours) & 1 step &  \textbf{1.43} & \textbf{76 / 112} & \textbf{2.1} \\
                \bottomrule
    \end{tabular}
     }
      \vspace{-3mm}
    \caption{\small{
    Desktop human perception study (112 rankings, 21 videos, 13 participants), with methods ranked from 1 (best) to 4 (worst). Our method achieved an average rank of 1.43, significantly outperforming others -- M2SVid ranked best $2.6\times$  more often than StereoCrafter (25 steps) and $4.75\times$  more often than SVG -- while being 6$\times$ and 635$\times$ faster, respectively. Runtimes were measured on an A100 GPU for $512\times512$ 16-frame videos.
        \textbf{
        \label{tab:desktop_study}
        }}}
    \vspace{-0.3cm}
\end{table} 

%% file: tables/headset_study.tex
\begin{table}[]
    \centering
    \resizebox{0.85\linewidth}{!}{
    \begin{tabular}{@{}lcc@{}}
    	\toprule
                 \multirow{2}{*}{VR headset comparison}  & StereoCrafter~\cite{shi2024stereocrafter} (25 steps)  & 
                 \cellcolor{almond} Ours (1 step) \\
                 
                  & preferred & \cellcolor{almond} preferred \\
                 \midrule 
                 StereoCrafter vs. Ours & 9 / 105 & \cellcolor{almond}\textbf{39 / 105}  \\
                \bottomrule
    \end{tabular}
     }
      \vspace{-3mm}
    \caption{\small\fix{{VR headset human perception study with 105 comparisons over 21 videos with 5 distinct users. Our method shows a clear advantage over StereoCrafter with 25 denoising steps. 
        \textbf{
        \label{tab:headset_study}
        }}}}
    \vspace{-0.5cm}
\end{table} 

%% file: figures/ablations.tex
\begin{figure}[b]
\centering
\vspace{-4mm}
\includegraphics[width=0.94\linewidth]{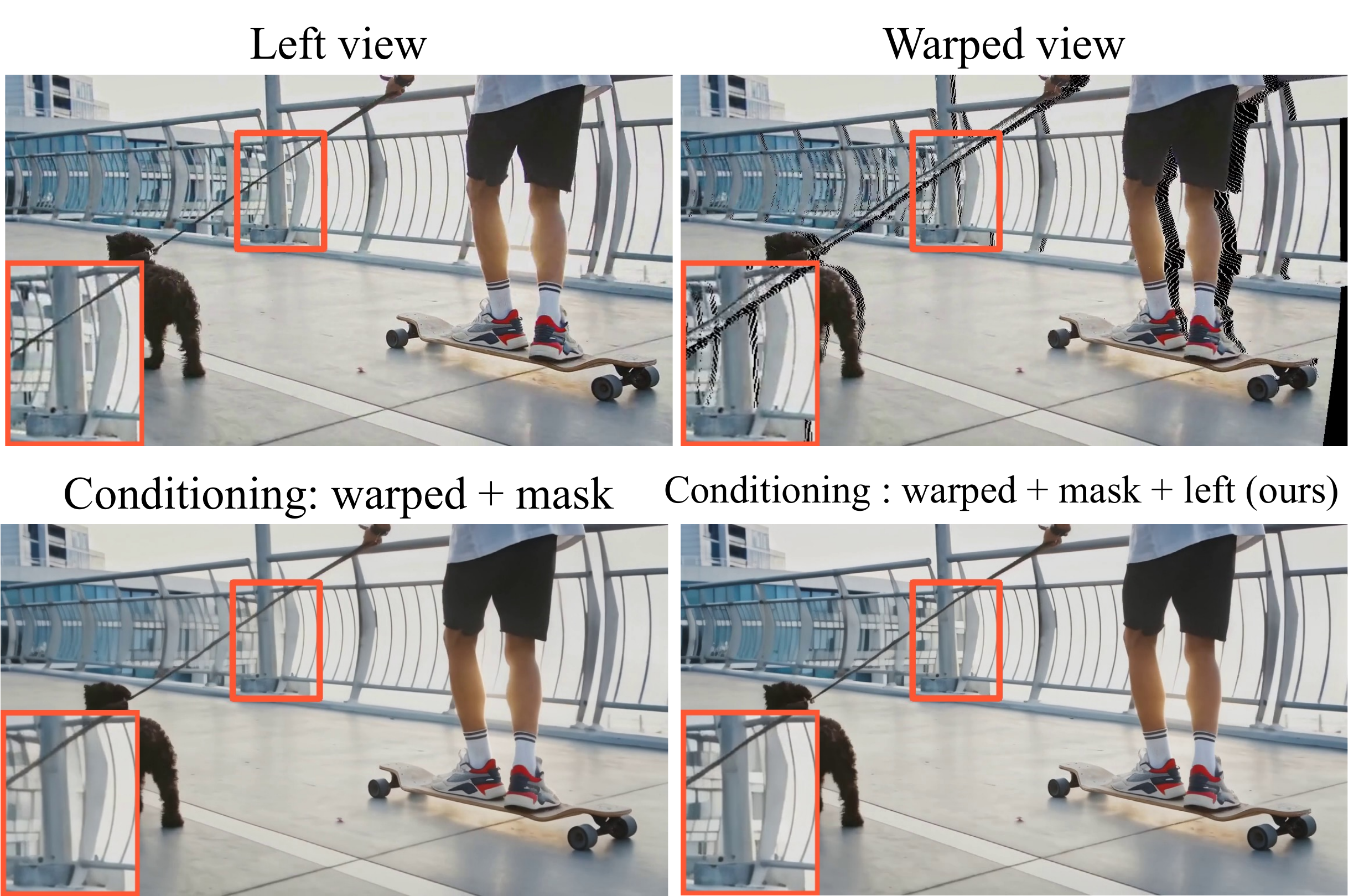}
\vspace{-3mm}
\caption{\small{Impact of our left view conditioning (Sec.~\ref{left-cond}).
        \textbf{
        \label{fig:alb_left_cond}
        }}}
\end{figure}

\begin{figure}[b]
\centering
\vspace{-3mm}
\includegraphics[width=0.97\linewidth]{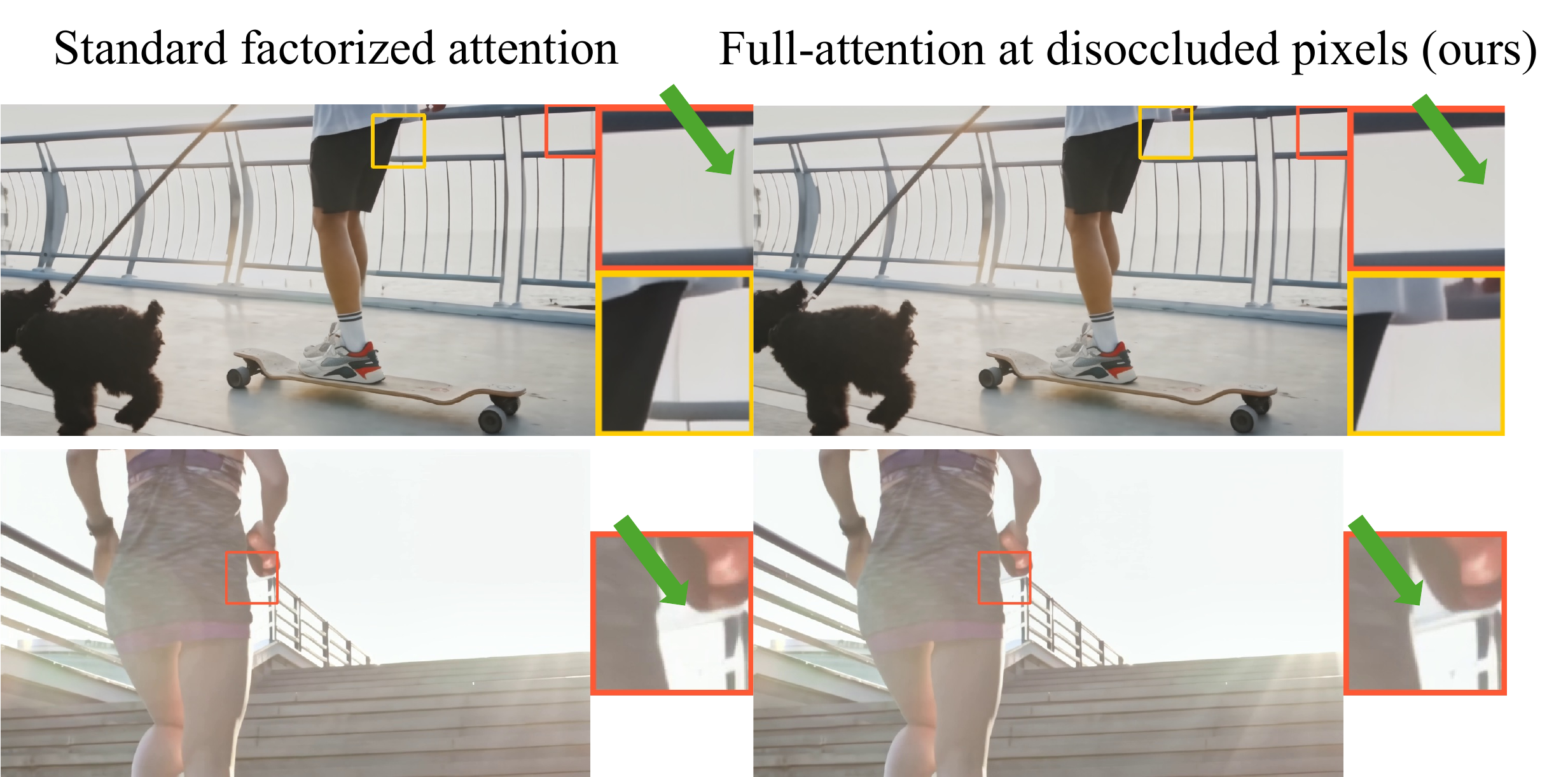}

\vspace{-3mm}
\caption{\small{Using full-attention at dis-occluded pixels (Ours, ~\cref{full-attention}) enables the model to exploit information from visible pixels in other frames to improve inpainting (see Appendix).
\label{fig:alb_full_attention}
}}
\end{figure}

%% file: tables/ablations/conditioning.tex
\begin{table}[]
    \setlength{\tabcolsep}{2pt}
    
    \centering
    \resizebox{\columnwidth}{!}{
    \begin{tabular}{@{}l@{~}cc|ccc|ccc@{}}
    	\toprule
                Model  & feed  &  Inf.  &  \multicolumn{3}{c}{\textbf{Stereo4D}} & \multicolumn{3}{c}{\textbf{Ego4D}} \\
                Architecture & forward & steps & PSNR$\uparrow$ &  MS-SSIM$\uparrow$ &  LPIPS$\downarrow$   & PSNR$\uparrow$ &  MS-SSIM$\uparrow$ &  LPIPS$\downarrow$   \\
                \midrule
                Cond. with $V^{warp}_r$ + $M^{occ}_r$ & \xmark & 25 & 
24.5 & 0.886 & 0.226 & 20.8 & 0.822 & 0.296 \\
                Cond. with $V^{warp}_r$ + $M^{occ}_r$ + $V_l$ & \xmark & 25 & 24.8 & 0.891 & 0.215 & 22.1 & 0.870 & 0.267 \\
                \midrule
                Cond. with $V^{warp}_r$ + $M^{occ}_r$ & \cmark & 1 & 26.1 & 0.913 & 0.187 & 21.5 & 0.837 & 0.276\\
                \rowcolor{almond} Cond. with $V^{warp}_r$ + $M^{occ}_r$ + $V_l$ & \cmark & 1 & 26.2 & 0.915 & 0.179 & 22.8 & 0.886 & 0.244\\
                \rowcolor{almond} +full attention & \cmark & 1 &26.2 & 0.915 & 0.180 & 22.7 & 0.885 & 0.248\\
                \bottomrule
    \end{tabular}
     }
      \vspace{-0.3cm}
    \caption{\small{Impact of left-view conditioning and full attention at disoccluded pixels on Stereo4D and Ego4D datasets.
        \textbf{
        \label{tab:abl_cond}
        }}}
    \vspace{-0.3cm}
\end{table}

%% file: tables/ablations/end_to_end.tex
\begin{table}[]
    \setlength{\columnwidth}{2pt}
    
    \centering
    \resizebox{1\linewidth}{!}{
    \begin{tabular}{@{}l@{~~}l@{~}@{~}cc|c@{~}c@{~}c|c@{~}c@{~}c@{}}
    	\toprule
           &      Loss & feed &  Inf.  & \multicolumn{3}{c}{\textbf{Stereo4D }} & \multicolumn{3}{c}{\textbf{Ego4D}}\\
              & &  forward &  steps  & PSNR$\uparrow$ &  MS-SSIM$\uparrow$ &  LPIPS$\downarrow$   & PSNR$\uparrow$ &  MS-SSIM$\uparrow$ &  LPIPS$\downarrow$   \\
                \midrule
          A &       Standard loss~\eqref{eq:diffusion_loss} & \xmark & 25 & 24.8 & 0.891 & 0.215 & 22.1 & 0.870 & 0.267 \\
            B &      Standard loss~\eqref{eq:diffusion_loss}  & \xmark & 1 & 25.7 & 0.901 & 0.242 & 23.0 & 0.877 & 0.320 \\
           C &        $L_{latent}$~\eqref{eq:latent} & \cmark & 1 &25.6 & 0.901 & 0.238 & 22.9 & 0.875 & 0.318 \\
      \rowcolor{almond}      D &       $L_{latent} + L_{\text{LPIPS}} + L_{\text{L1}}$~\eqref{eq:training_loss} & \cmark & 1 & 26.2 & 0.915 & 0.179 & 22.8 & 0.886 & 0.244 \\
      
                \bottomrule
    \end{tabular}
     }
      \vspace{-0.3cm}
    \caption{\small{Impact of different inference strategies and losses. 
        \textbf{
        \label{tab:abl_end_to_end}
        }}}
    \vspace{-5mm}
\end{table}

%% file: sec/5_conclusion.tex
\section{Conclusion}
We introduce an end-to-end approach for stereoscopic video inpainting and refinement in this work. 
First, we extend the SVD model to take the input left video, warped right video, and disocclusion mask as conditioning.
Next, we modify the attention layers in SVD to compute full attention for the discoccluded pixels in order to improve inpainting quality. Crucially, we perform the video refinement using a single denoising step, enabling end-to-end training with image space losses. Qualitative and quantitative experiments, as well as user studies, demonstrate that our method clearly outperforms prior state-of-the-art methods for the monocular-to-stereo video conversion task.

%% file: sec/supplementary.tex
\appendix
The supplementary material provides more details about the datasets, implementation, and results. \cref{sec:diffusion_background_sup} provides a brief background on Diffusion Models. More details about our training and evaluation datasets are provided in \cref{sec:dataset_details_sup}. \cref{sec:implementation_details_sup} provides further implementation details as well as additional information on user studies, while \cref{sec:details_on_results_sup} contains more detailed results and analysis. \cref{sec:high_resolution} provides details on inference for high-resolution long videos. Finally, \cref{sec:limitations} discusses the limitations of our approach. 

\section{Background on Diffusion Models}
\label{sec:diffusion_background_sup}

\parsection{DDPMs} Denoising Diffusion Probabilistic Models (DDPMs)~\cite{ho2020denoising} are generative models trained to map a simple noise distribution $p_T$ to the data distribution $p_0$, by reversing a stochastic forward process $p_t$, $t = 1, ..., T$. This forward process gradually adds small amounts of Gaussian noise with variance $\beta_t$, ensuring the reverse process can be approximated as a Gaussian distribution. The forward process might be also defined by $\vect{x}_t = \sqrt{\bar\alpha_t}\vect{x}_0 + \sqrt{1 - \bar\alpha_t} \vect{\epsilon}$, where $\vect{x}_0$ is a data sample, $\vect{\epsilon} \sim \mathcal{N}(\text{0}, \text{I})$ is a noise, and $\alpha_t = 1 - \beta_t$ and $\bar\alpha_t = \prod_{\tau=1}^t \alpha_\tau$. A denoising model $\hat{\vect{v}}_{\theta}(x_t, t)$ is trained to progressively remove noise from the input, by predicting $\vect{x}_{t-1}$ from $\vect{x}_t$ and timestamp $t$, effectively reversing the diffusion.

\parsection{DDIMs} Denoising Diffusion Implicit Models (DDIMs)~\cite{song2020denoising} extend DDPMs by introducing non-Markovian diffusion processes, preserving the original training objective while enabling much faster inference with as few as 25 or 50 steps.

\parsection{LDMs} To facilitate training, Latent Diffusion Models (LDMs)~\cite{rombach2022high} operate in the latent space of a Variational Autoencoder (VAE)~\cite{kingma2013auto}, which consists of an encoder $E$ and a decoder $D$. The VAE is trained separately and remains frozen during diffusion model training. By working in the latent space, where $D(E(x)) \approx x$, LDMs significantly reduce computational complexity while preserving essential data structure. The diffusion model $\hat{\vect{v}}_{\theta}$ is then trained directly in this lower-dimensional latent space.

\parsection{Conditional diffusion models} Finally, conditional diffusion models~\cite{saharia2022photorealistic, zhang2023adding} extend diffusion models by conditioning the denoising model $\hat{\vect{v}}_{\theta}(x_t, t, c)$ on additional input $c$, such as text~\cite{saharia2022photorealistic}, images~\cite{saharia2022photorealistic}, or pose and depth maps~\cite{zhang2023adding}.

\begin{figure*}
\centering
\includegraphics[width=0.95\linewidth]{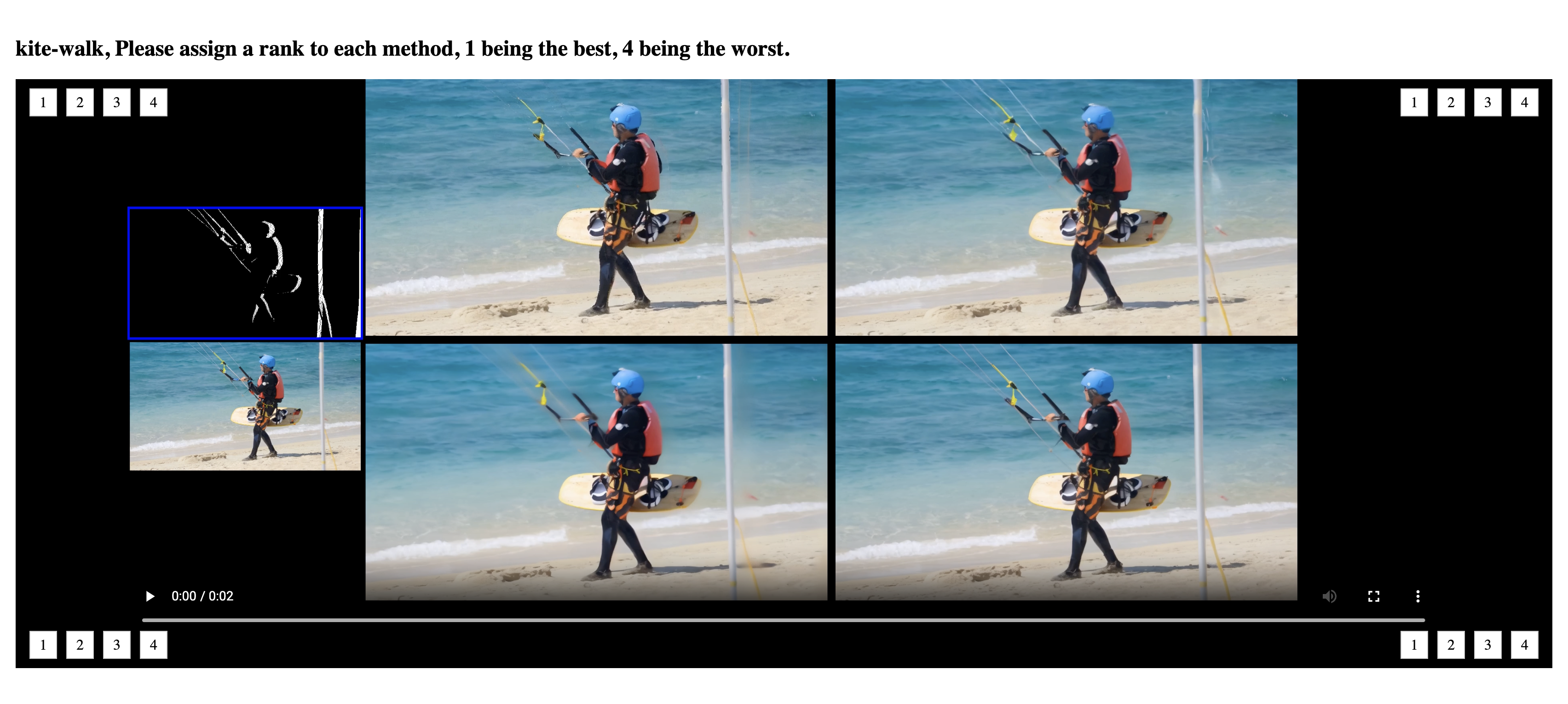}
\vspace{-3mm}
\caption{\small{The interface shown to the users during the human perception study. On the left, we showed the input left video, together with the disocclusion mask. For each video, we showed results of the four method. Note that the users were not shown any labels to indicate which outputs were generated by which method. Furthermore, we randomized the order in which the methods were listed for each video. The users were asked to assign rankings from 1 (best) to 4 (worst) for each method. In cases two or more methods were indistinguishable, the users were given the choice of assigning the same rank to multiple methods. The users could play or pause the videos as they wish, view the video in full screen, or change the playback speed if they desired.
        \textbf{
        \label{fig:sup_user_study_interface}
        }}}
\end{figure*}

\begin{figure*}
\centering
\includegraphics[width=0.95\linewidth]{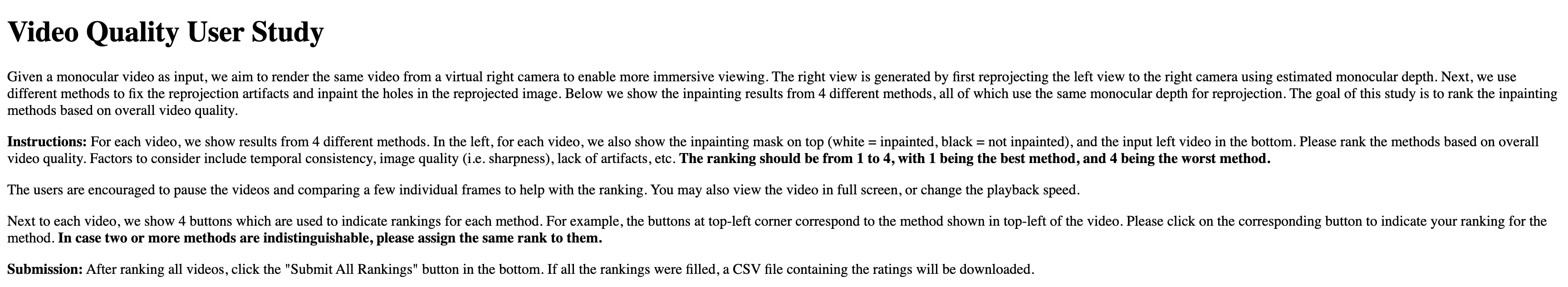}
\vspace{-3mm}
\caption{\small{The users were shown the above intructions during the study.
        \textbf{
        \label{fig:sup_user_study_instructions}
        }}}
\end{figure*}

\section{More Details on Datasets}
\label{sec:dataset_details_sup}
\subsection{Stereo4D}
\noindent\textbf{Stereo4D:} The dataset contains $\sim200$K video clips sourced from $\sim7.5$K  online videos. Each clip lasts between 2–6 seconds with a spatial resolution of $512\times512$. The videos cover diverse scenes featuring both indoor and outdoor activities. While the dataset is relatively large, it also includes many static videos or videos with low disparity. 
We hold out a random subset of 400 videos for method evaluation and use the rest for the training. We source only one clip per video that is longer than 3.2 seconds and sample 16 frames at 5 fps, resulting in a test set of 400 diverse video clips with a resolution of $16\times512\times512$.

\subsection{Ego4D}
\noindent\textbf{Ego4D:} This is one of the largest egocentric video datasets, where participants were asked to perform various activities. While most videos are monocular, the dataset includes 263 long stereoscopic videos (80 hours in total) 
with a resolution of $1400\times1400$. As the videos are egocentric, captured using a head-mounted camera, they exhibit significant camera movement and contain close objects with a large disparity range between the stereo views. 
We split 263 stereo videos of the dataset into $\sim$ 57k 5-second video clips. We hold out all clips from one video, namely 200 clips, for method evaluation, while using the rest 57K clips for training.

\parsection{Ego4d Pre-processing}
We perform rectification for each long video as described in Sec. 5.5 of the main paper. For video rectification, we uniformly sample 200 frames from each video. After rectifying the videos, we spatially crop the video to discard boundary regions with missing pixels. Then, we split the videos into 5-second subsequent video clips and compute shifts and disparity at the clip level. Finally, to ensure high-quality rectified data, we perform LoFTR feature matching~\cite{sun2021loftr} again and filter point pairs by computing the fundamental matrix with RANSAC~\cite{fischler1981random}, applied to frames sampled at 1 FPS from each clip. We then filter out clips where the matched points exhibit more than 2 pixels of vertical disparity.

\subsection{DAVIS}

\noindent\textbf{DAVIS:} The dataset consists of 50 videos, each up to 6 seconds long, with a resolution of $1024\times1920$. The videos are highly dynamic, featuring moving cameras and one or more moving objects (e.g., BMX riding, breakdancing, etc.). Note that the dataset only contains monocular videos, without a ground truth depth or ground truth right view. Hence we use this dataset solely for qualitative comparisons and user studies.

\begin{table*}[]
    
    \centering
    \resizebox{1\linewidth}{!}{
    \begin{tabular}{@{}lcc|ccc |ccc | ccc@{}}
    	\toprule
                Method & Training data & Denoising steps & \multicolumn{3}{c}{Full view} & \multicolumn{3}{c}{Inside disocclusion mask only} & \multicolumn{3}{c}{Outside disocclusion mask}  \\
                & & & PSNR$\uparrow$ &  MS-SSIM$\uparrow$ &  LPIPS$\downarrow$   & PSNR$\uparrow$ &  MS-SSIM$\uparrow$ &  LPIPS$\downarrow$   & PSNR$\uparrow$ &  MS-SSIM$\uparrow$ &  LPIPS$\downarrow$ \\
                \midrule
                SVG~\cite{dai2024svg} & - (training free) & 50 steps & 25.6 & 0.926 & 0.217 & 38.1 & 0.994 & 0.014 & 26.3 & 0.940 & 0.190 \\
                StereoCrafter~\cite{shi2024stereocrafter} & private dataset  & 25 steps &24.9 & 0.909 & 0.242 & 38.2 & 0.995 & 0.012 & 25.5 & 0.922 & 0.217 \\
                StereoCrafter~\cite{shi2024stereocrafter} & private dataset  &1 step & 25.3 & 0.911 & 0.262 & 38.9 & 0.996 & 0.014 & 25.8 & 0.923 & 0.234 \\
               \rowcolor{almond} \ours (Ours) & Stereo4D + Ego4D & 1 step & 26.2 & 0.915 & 0.180 &38.2 & 0.994 & 0.010 & 26.8 & 0.924 & 0.161 \\
               
                \bottomrule
    \end{tabular}
     }
      \vspace{-0.2cm}
    \caption{\small{Quantitative comparison of our approach with state-of-the-art methods on Stereo4D test set, in terms of PSNR, MS-SSIM, and LPIPS. We report the metrics computed over the full image, only inside the discoccluion mask, and only outside the disocclusion mask. 
        \textbf{
        \label{tab:sota-full}
        }}}
\end{table*}

\begin{table*}[]
    \setlength{\tabcolsep}{2pt}
    
    \centering
    \resizebox{1\linewidth}{!}{
    \begin{tabular}{@{}lcc|ccc|ccc|ccc|ccc@{}}
    	\toprule
                Model  & End-to-end  &  Denoising  &
                \multicolumn{6}{c}{Full image} & \multicolumn{6}{c}{Inside disoccluded regions} \\
               Architecture & training & steps &  \multicolumn{3}{c}{Stereo4D} & \multicolumn{3}{c}{Ego4D} &  \multicolumn{3}{c}{Stereo4D} & \multicolumn{3}{c}{Ego4D} \\
                & & & PSNR$\uparrow$ &  MS-SSIM$\uparrow$ &  LPIPS$\downarrow$   & PSNR$\uparrow$ &  MS-SSIM$\uparrow$ &  LPIPS$\downarrow$   & PSNR$\uparrow$ &  MS-SSIM$\uparrow$ &  LPIPS$\downarrow$ & PSNR$\uparrow$ &  MS-SSIM$\uparrow$ &  LPIPS$\downarrow$ \\
                \midrule
                Cond. with $V^{warp}_r$ + $M^{occ}_r$ & \xmark & 25 steps & 
24.5 & 0.886 & 0.226 & 20.8 & 0.822 & 0.296 & 36.1 & 0.993 & 0.011 & 29.3 & 0.986 & 0.027 \\
                Cond. with $V^{warp}_r$ + $M^{occ}_r$ + $V_l$ & \xmark & 25 steps & 24.8 & 0.891 & 0.215 & 22.1 & 0.870 & 0.267 & 36.7 & 0.994 & 0.011 & 30.1 & 0.988 & 0.024 \\
                \midrule
                Cond. with $V^{warp}_r$ + $M^{occ}_r$ & \cmark & 1 step & 26.1 & 0.913 & 0.187 & 21.5 & 0.837 & 0.276 & 38.0 & 0.994 & 0.010 & 30.4 & 0.987 & 0.026 \\
                \rowcolor{almond} Cond. with $V^{warp}_r$ + $M^{occ}_r$ + $V_l$ & \cmark & 1 step & 26.2 & 0.915 & 0.179 & 22.8 & 0.886 & 0.244 & 38.2 & 0.995 & 0.009 & 30.9 & 0.989 & 0.024 \\
                \rowcolor{almond} +full attention & \cmark & 1 step &26.2 & 0.915 & 0.180 & 22.7 & 0.885 & 0.248 & 38.2 & 0.994 & 0.010 & 30.7 & 0.989 & 0.025 \\
                \bottomrule
    \end{tabular}
     }
      \vspace{-0.2cm}
    \caption{\small{We analyse the impact of different conditioning inputs as well as inference modes on Stereo4D and Ego4D datasets. The metrics are reported over the full images, as well as only the disoccluded regions.
        \textbf{
        \label{tab:full_abl_cond}
        }}}
\end{table*}

\begin{table*}[]
    \setlength{\tabcolsep}{2pt}
    
    \centering
    \resizebox{1\linewidth}{!}{
    \begin{tabular}{@{}llcc|ccc|ccc|ccc|ccc@{}}
    	\toprule
           &      Loss & End-to-end &  Inference  &
                \multicolumn{6}{c}{Full image} & \multicolumn{6}{c}{ Inside disoccluded regions} \\
                 &  training &  steps & &  \multicolumn{3}{c}{Stereo4D} & \multicolumn{3}{c}{Ego4D} &  \multicolumn{3}{c}{Stereo4D } & \multicolumn{3}{c}{Ego4D} \\
              &   & & & PSNR$\uparrow$ &  MS-SSIM$\uparrow$ &  LPIPS$\downarrow$   & PSNR$\uparrow$ &  MS-SSIM$\uparrow$ &  LPIPS$\downarrow$   & PSNR$\uparrow$ &  MS-SSIM$\uparrow$ &  LPIPS$\downarrow$ & PSNR$\uparrow$ &  MS-SSIM$\uparrow$ &  LPIPS$\downarrow$ \\
                \midrule
          A &       Standard diffusion loss & \xmark & 25 steps & 24.8 & 0.891 & 0.215 & 22.1 & 0.870 & 0.267 & 36.7 & 0.994 & 0.011 & 30.1 & 0.988 & 0.024 \\
            B &      Standard diffusion loss  & \xmark & 1 step & 25.7 & 0.901 & 0.242 & 23.0 & 0.877 & 0.320 & 38.5 & 0.995 & 0.012 & 31.5 & 0.990 & 0.030 \\
           C &        $L_{latent}$ & \cmark & 1 step &25.6 & 0.901 & 0.238 & 22.9 & 0.875 & 0.318 & 38.3 & 0.995 & 0.012 & 31.4 & 0.990 & 0.030 \\
      \rowcolor{almond}      D &       $L_{latent} + L_{\text{LPIPS}} + L_{\text{L1}}$ & \cmark & 1 step & 26.2 & 0.915 & 0.179 & 22.8 & 0.886 & 0.244 & 38.2 & 0.995 & 0.009 & 30.9 & 0.989 & 0.024 \\

                \bottomrule
    \end{tabular}
     }
      \vspace{-0.2cm}
    \caption{\small{We analyse the impact of different training strategies and losses on the Stereo4D and Ego4D datasets. Metrics are reports over the full image, as well as over only the disoccluded regions.
        \textbf{
        \label{tab:abl_full_end_to_end}
        }}}
    \vspace{-0.5cm}
\end{table*}

\section{Implementation Details}
\label{sec:implementation_details_sup}
\subsection{Conditioning}

We extend the first convolution of SVD that initially took 8-dimensional input (4 dim noise latent + 4 dim VAE encoded image) to the 13-dimensional input (4 dim noise latent + 4 dim VAE encoded left to view video + 4 dim VAE encoded warped view video + 1 dim disocclusion mask).  To ensure smooth model initialization we copy weights 5-8 channels to 9-12 channels and divide both by factor 2 and initialize channels of dissolution mask with 0.

\parsection{Model training} We initialize our M2SVid model using the Stable Video Diffusion model (\texttt{stable-video-diffusion-img2vid-xt} version)~\cite{blattmann2023stable}. We train the model for 300k iterations with a batch size of 16 and a learning rate of $2 \times 10^{-6}$. Due to GPU memory limitations, during training, we sample batches with frame number and resolutions of $4\times512\times512$, $16\times256\times256$, and $25\times192\times192$, utilize gradient checkpointing\cite{chen2016training} and train the model in float16 precision. 
We employ random resized cropping, using a resize scale in the range of [0.3, 1.0], for data augmentation. We preprocess the disocclusion masks using the closing morphological transformation to fill small holes inside the inpainting regions with a kernel size of 11 during training and testing.

\parsection{Depth estimation with DepthCrafter}
During inference for depth prediction, we utilize the DepthCrafter~\cite{hu2024depthcrafter} method, the state-of-the-art in diffusion-based video-depth estimation approach. This model can accommodate up to 110 frames and output temporally consistent depth. However, any depth model might be used at this step, for example, efficient 1-step depth diffusion models as in~\cite{garcia2024fine}. 
We predict depth in chunks of 110 frames with an overlap of 25 frames. Depth was predicted on frames resized to a resolution of 1024, followed by upscaling to the original size. 

\subsection{Desktop User Study}
In order to perform our desktop human perception user study to rank the different methods, we selected a set of 21 videos. We tried to ensure a diverse collection of videos covering different types of subjects, video quality, and motion pattern. We avoided including without with very large motions since we found it very hard to compare the methods in such cases.

During the study, the participants were provided HTML pages containing the links to the generated predictions, as shown in Figure~\ref{fig:sup_user_study_interface}.  Furthermore, the users were shown detailed instructions about the study at the top of the HTML page. The exact instructions are shown in Figure~\ref{fig:sup_user_study_instructions}. In particular, the users were provided a brief background about the problem statement and the method. They were asked to rank the quality of the different methods based on  temporal consistency, image quality (i.e. sharpness), lack of artifacts, \etc. In order to miminize noise, the users were allowed to assign same ranks to two or more methods in case they were indistinguishable. 

We had 13 distinct participants in the study, out of which 9 were male and 4 were female. We divided the 21 videos into sets of 7 each. Each user was given the option of assigning rankings to the videos in one or two sets each. In total, we received rankings for 16 sets, resulting in 112 sets of ranks.

\subsection{VR Headset User Study}

For the VR headset user study, the participants were shown the stereoscopic videos generated by our method and StereoCrafter (25 denoising steps) on a video player that supports side-by-side (3D) content. We used the same 21 videos that were used in the Desktop user study. The participants were then asked to rank the methods based on general viewing experience, while allowing for tied ratings. Note that all the videos were anonymized and the order of showing the methods was also randomized. We had 5 distinct participants in the user study (3 male and 2 female), each of whom provided rankings for all 21 videos.

\section{More Details on Results}
\label{sec:details_on_results_sup}
\subsection{State-of-the-art Methods}

\textbf{SVG} is a training-free method that utilizes a frozen diffusion model to inpaint regions within the mask while preserving the other regions. To achieve spatial-temporal consistency, the method employs inpainting of 8 uniformly sampled views between the given left view and the required right view at the same time, increasing the computational burden. We use 50 denoising steps as recommended. SVG~\cite{dai2024svg} has been compared to state-of-the-art video inpainting~\cite{zhou2023propainter, li2022towards} and dynamic novel view synthesis~\cite{liu2023robust, li2023dynibar} methods, significantly outperforming them in temporal consistency and overall inpainting quality. This highlights the fundamental differences between stereo video inpainting, classical video inpainting, and novel view synthesis, revealing a substantial gap between these tasks. Due to this disparity, we exclude classical video inpainting and novel view synthesis methods from our comparisons. 

\textbf{StereoCrafter}~\cite{shi2024stereocrafter} is a concurrent method and the closest baseline to our approach. It fine-tunes a diffusion model, SVD, to perform inpainting given a warped view and a mask in a diffusion-based denoising setup, trained on a privately collected internet dataset. Since the method does not specify the number of denoising steps, we use 25 steps, as is commonly done~\cite{hu2024depthcrafter}.

Note that our method \ours was trained on the Stereo4D and Ego4d datasets, while SVG is a training-free method, and StereoCrafter is trained on a private dataset. However, this StereoCrafter private dataset also contains general internet videos and should have a similar data distribution. Comparing zero-shot performance on another dataset is challenging, as there are no publicly available stereo video datasets in the general domain. For this reason, we evaluate on the Stereo4D dataset.

\subsection{State-of-the-art Comparison}

For quantitative evaluation, we use PSNR, MS-SSIM, and LPIPS~\cite{zhang2018perceptual} metrics, computed independently for each video and averaged over the full dataset. Additionally, we report these metrics separately for the disoccluded and non-disoccluded regions in~\cref{tab:sota-full}. To achieve this, we mask all pixels outside the considered region with white pixels, compute the metrics at the video level, and then average them over the dataset.

Our method obtains the best PSNR and LPIPS scores when averaged over the full image. For the disoccluded regions, we see that single step StereoCrafter obtains the best PSNR score, while also having the worse LPIPs score. We believe this is because the method generates blurry results, which is often favored by PSNR, but strongly penalized by LPIPS. Outside the discocclusion region, our method obtains the best LPIPS and PSNR scores while SVG obtains the best MS-SSIM score.

\subsection{Full-attention at Disoccluded Pixels.}

In~\cref{fig:alb_full_attention_1,fig:alb_full_attention_2,fig:alb_full_attention_3}, we provide further qualitative evaluation of our proposed full attention at disoccluded pixels. We find that full attention is particularly beneficial for scenes with complex backgrounds and strong camera movement, where both foreground and background pixels move and correct inpainting requires the model to ``copy'' disoccluded pixels from different spatial locations in other frames. For example, in~\cref{fig:alb_full_attention_1}, the handrail of the stairs in frame 4 can only be correctly inpainted by using information from a different spatial location in frame 9 due to camera movement. In~\cref{fig:alb_full_attention_2}, due to the strong movement of the motorcyclist behind the central rider, to inpaint the region near the helmet in frame 11, the model has to use data from a strongly shifted spatial location in frame 10. Finally, in~\cref{fig:alb_full_attention_3}, while standard factorized attention hallucinated the reins in frame 16 at the same spatial location as in the previous frame, creating a visual artifact as if the reins split into two pieces, full attention at disoccluded pixels prevents this hallucination and correctly inpaints the region.

\subsection{Ablations}

In~\cref{tab:full_abl_cond,tab:abl_full_end_to_end}, we further report metrics for the disoccluded regions for our ablation studies. We find that incorporating the left view as an additional conditioning signal (\cref{tab:full_abl_cond}) results in a slight improvement in inpainting performance across both datasets. Additionally, end-to-end training with latent space supervision (\cref{tab:abl_full_end_to_end}) achieves nearly the same inpainting performance as a model trained with the standard diffusion paradigm and evaluated with 1-step inference. However, end-to-end training with image-based loss leads to a substantial drop in LPIPS, particularly on the Ego4D dataset.

\subsection{Run-time}

\begin{table}[t]
    
    \centering
    \resizebox{1\linewidth}{!}{
    \begin{tabular}{@{}l|c@{}}
    	\toprule
                Method & Latency (s) $\downarrow$\\
                \midrule
                without full attention on the disoccluded regions& \textbf{2.0} \\
                with full attention on the disoccluded regions & \textbf{2.1} \\
                \bottomrule
    \end{tabular}
     }
      \vspace{-2mm}
    \caption{\small{The latency of two variations of our model: with and without full attention on the disoccluded regions. Enabling full attention on the disoccluded regions results in only a slight increase in runtime. The run-times computed
using an A100 GPU on a $512\times512$ videos with 16 frames.
        \textbf{
        \label{tab:run_time_fa}
        }}}
\end{table} 

\begin{table}[t]
\setlength{\columnsep}{4pt}%
\resizebox{1\columnwidth}{!}{
    \setlength{\columnwidth}{1pt}
    \begin{tabular}{@{}lcc|c|cc@{}}
    	\toprule
                  Training & \multirow{2}{*}{Loss} & End-to-end  &  Inference  & \multicolumn{2}{c}{LPIPS$\downarrow$}   \\
                 regime & & training & steps & Stereo4D & Ego4d \\
                 
                 \midrule
                \multirow{3}{*}{Sampling $t$} & \multirow{3}{*}{$L_{latent}$} & \multirow{3}{*}{\xmark} &  1 ($t=T$) & 0.242 & 0.320 \\
                & & & 5 & 0.217 & 0.278 \\
                & & & 25 & 0.215 & 0.267 \\
                
                \midrule
                \multirow{3}{*}{Fixed $t=T$} & \multirow{3}{*}{$L_{latent} + L_{\text{LPIPS}} + L_{\text{L1}}$ } & \multirow{3}{*}{\cmark} &  \cellcolor{almond} 1 ($t=T$) & \cellcolor{almond} \textbf{0.179} &  \cellcolor{almond} \textbf{0.244}  \\
                & & & 5 & 0.200 & 0.260  \\
                & & & 25 & 0.243 & 0.298  \\
                \bottomrule
    \end{tabular}
     }
     \caption{\small{\fix{The effect of the number of inference steps on model performance. In standard training with sampled $t$, more inference steps improve results, while 1-step inference performs poorly. In contrast, our model is trained with fixed $t=T$, enabling end-to-end supervision in image space (which is not possible with sampled $t$). This leads to superior 1-step performance, while multi-step inference degrades results due to mismatch with training.}
        \textbf{
        \label{tab:inf_steps}
        }}}
\end{table}

\begin{figure*}[t!]
\centering
\begin{subfigure}{0.245\linewidth}
\centering
    \includegraphics[width=1\linewidth]{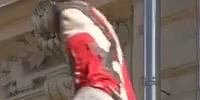}
    \caption{ Crop from input 
    left view\label{fig:blurriness:left}}
\end{subfigure}%
\hspace{1.mm}%
\begin{subfigure}{0.245\linewidth}
\centering
    \includegraphics[width=1\linewidth]{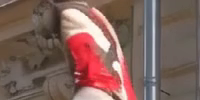}
    \caption{ VAE encoded-decoded left view\label{fig:blurriness:rec}}
\end{subfigure}%
\hspace{1.mm}%
\begin{subfigure}{0.245\linewidth}
\centering
    \includegraphics[width=1\linewidth]{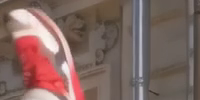}
    \caption{ Generated right (stitching area) \label{fig:blurriness:stitching}}
\end{subfigure}
\hspace{0.5mm}%
\begin{subfigure}{0.245\linewidth}
\centering
    \includegraphics[width=1\linewidth]{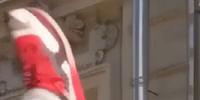}
    \caption{ Generated right (not stitching area)\label{fig:blurriness:nostitching}}
\end{subfigure}
\vspace{-3.3mm}
\caption{Blurriness due to usage of VAE and spatial stitching in high-resolution videos.}
\label{fig:blurriness}
\vspace{-3.3mm}
\end{figure*}

\input{figures/supmat_full_attention}

We compare the runtime of our model with and without full attention in the disoccluded regions in \cref{tab:run_time_fa}. Our results show that enabling full attention leads to only a slight increase in runtime.

\subsection{\fix{Number of Denoising Steps}}

\fix{In~\cref{tab:inf_steps}, we analyze the effect of the number of inference steps on model performance. In a standard training regime, where a timestamp $t$ is randomly sampled during training, one-step inference results in poor performance, while increasing the number of inference steps improves it.  
In contrast, we explicitly train a one-step model by fixing the timestamp to $t = T$ during training. This enables end-to-end training with an image-space loss, as the U-Net directly predicts denoised latents that can be decoded with a VAE and supervised in image space. Note that training with an image-space loss is not possible in the standard regime with sampled $t$, since when $t \ne T$, the U-Net outputs a mixture of latents and noise (not clean latents), which therefore cannot be decoded with the VAE and supervised in image space.  
As a result of end-to-end training with a single step, our model achieves the best performance with one-step inference. However, multi-step inference degrades performance, as the model was not trained with $t \ne T$.}

\section{Inference on Arbitrary Length and Resolution }
\label{sec:high_resolution}
To achieve stereo video conversion for videos of arbitrary length, we fine-tune our M2SVid model for 20K iterations using the auto-regressive modeling approach proposed in~\cite{shi2024stereocrafter}. Specifically, during training, we replace the first $n$ frames of $V_{r}^{warp}$ with ground truth frames from the right video $V_{r}$, where $n$ is randomly sampled between $0$ and $N$. During inference on long videos, we use the last $m$ frames generated in the current round as input for the next round, ensuring seamless stitching between consecutive rounds. For inference with arbitrary resolution, following~\cite{shi2024stereocrafter}, we utilize tiled diffusion. Specifically, we divide high-resolution videos into overlapping blocks and perform stereo video conversion on each block independently. The overlapping regions are blended in the latent space before VAE decoding to ensure seamless transitions between blocks. We provide examples generated by our model on high-resolution long videos in the supplementary material zip file.

\section{\fix{Limitations}}
\label{sec:limitations}

\fix{
We build our model upon the Stable Video Diffusion model, which allows us to leverage learned video priors obtained through large-scale generative pretraining. Due to the high computational cost of processing video data, SVD employs a VAE to compress videos into a lower-resolution latent space, where denoising is performed. Following this architecture, we first encode the video using a VAE, then apply a U-Net, and finally decode the output back to video space using the same VAE. However, VAE compression may lead to loss of high-frequency details. As illustrated in~\cref{fig:blurriness}, even encoding and decoding the left-view videos with the VAE alone can introduce visible blurriness (e.g., the shoe in~\cref{fig:blurriness:left,fig:blurriness:rec}) and loss of fine details. To support stereo conversion for higher-resolution videos, we additionally apply temporal and spatial stitching, as described in~\cref{sec:high_resolution}. While simple and efficient, stitching may introduce further blurring artifacts, as shown in~\cref{fig:blurriness:stitching,fig:blurriness:nostitching}. We plan to address these limitations in future research.
}

%% file: figures/supmat_full_attention.tex
\begin{figure*}[b]
\centering
\vspace{-3mm}
\includegraphics[width=1\linewidth]{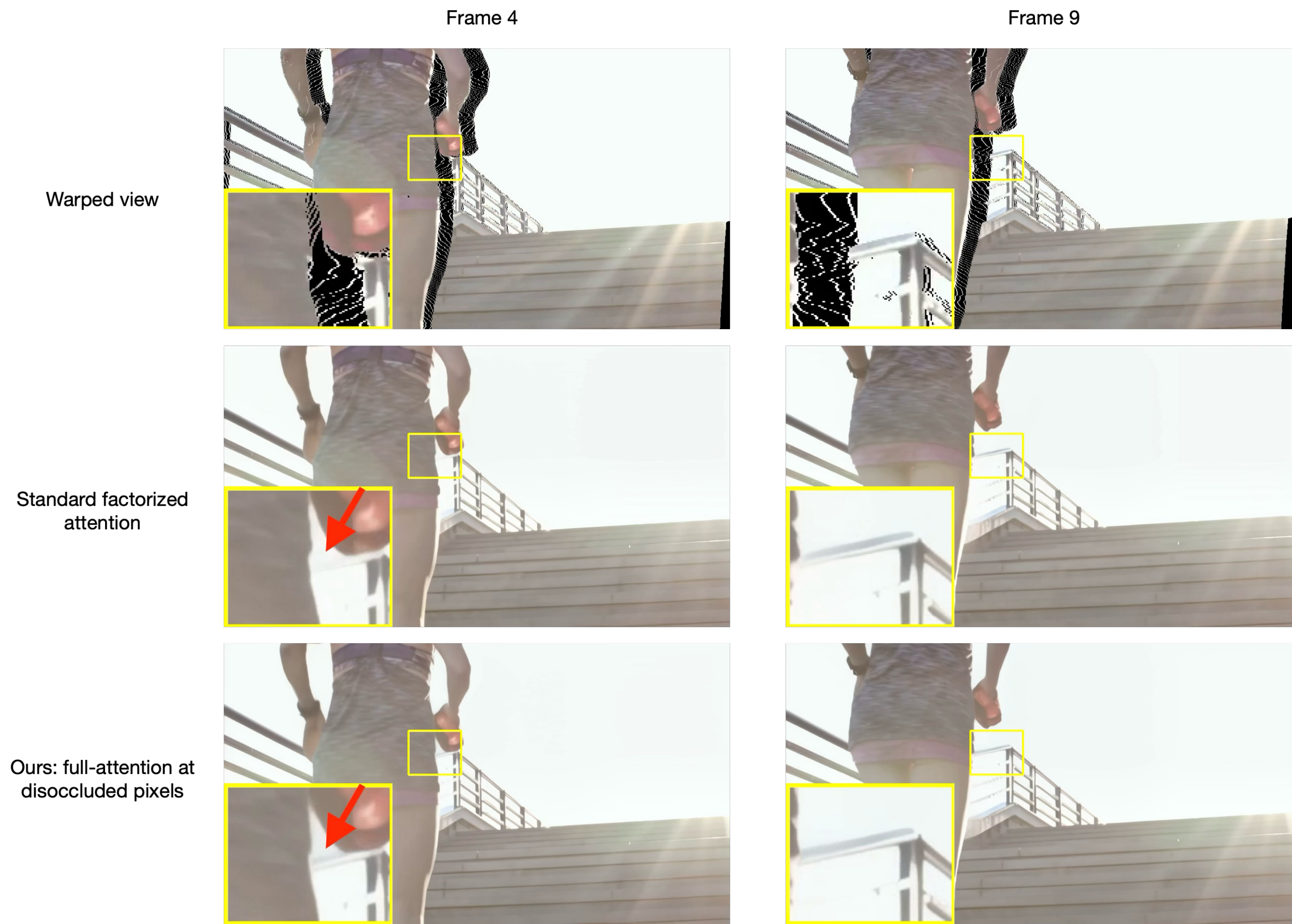}
\vspace{-2mm}
\caption{\small{Standard factorized attention vs full-attention at dis-occluded pixels (ours). Full attention at dis-occluded pixels helps the model better exploit information from other frames, in particular by correctly inpainting the handrail of the stairs using data from a different spatial location in another frame.
\label{fig:alb_full_attention_1}
}}
\end{figure*}

\begin{figure*}[b]
\centering
\vspace{-3mm}
\includegraphics[width=1\linewidth]{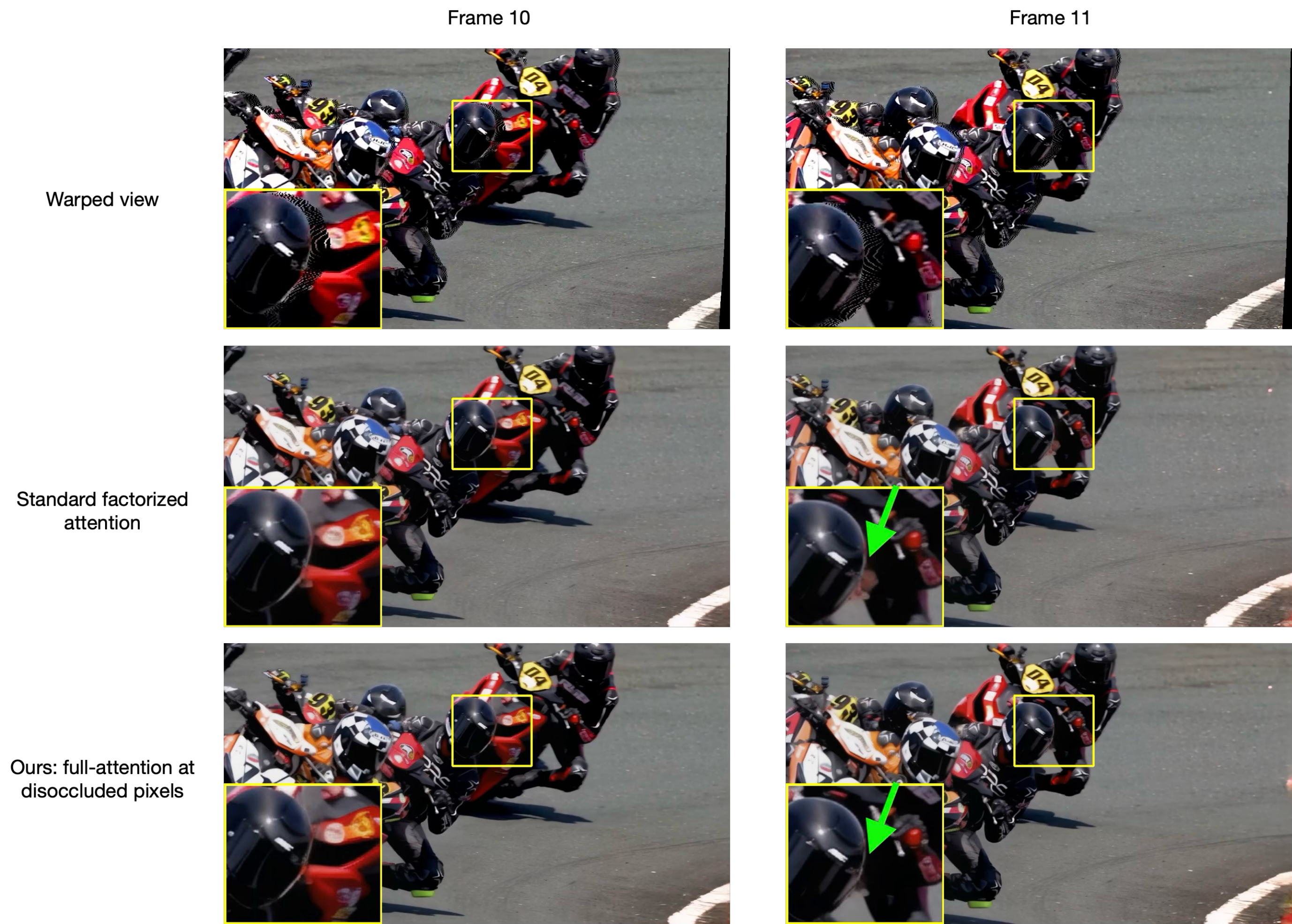}
\vspace{-2mm}
\caption{\small{Standard factorized attention vs full-attention at dis-occluded pixels (ours). Full attention at dis-occluded pixels helps the model better exploit information from other frames, in particular by correctly inpainting the region near the helmet, using data from a different spatial location in another frame.
\label{fig:alb_full_attention_2}
}}
\end{figure*}

\begin{figure*}[b]
\centering
\vspace{-3mm}
\includegraphics[width=1\linewidth]{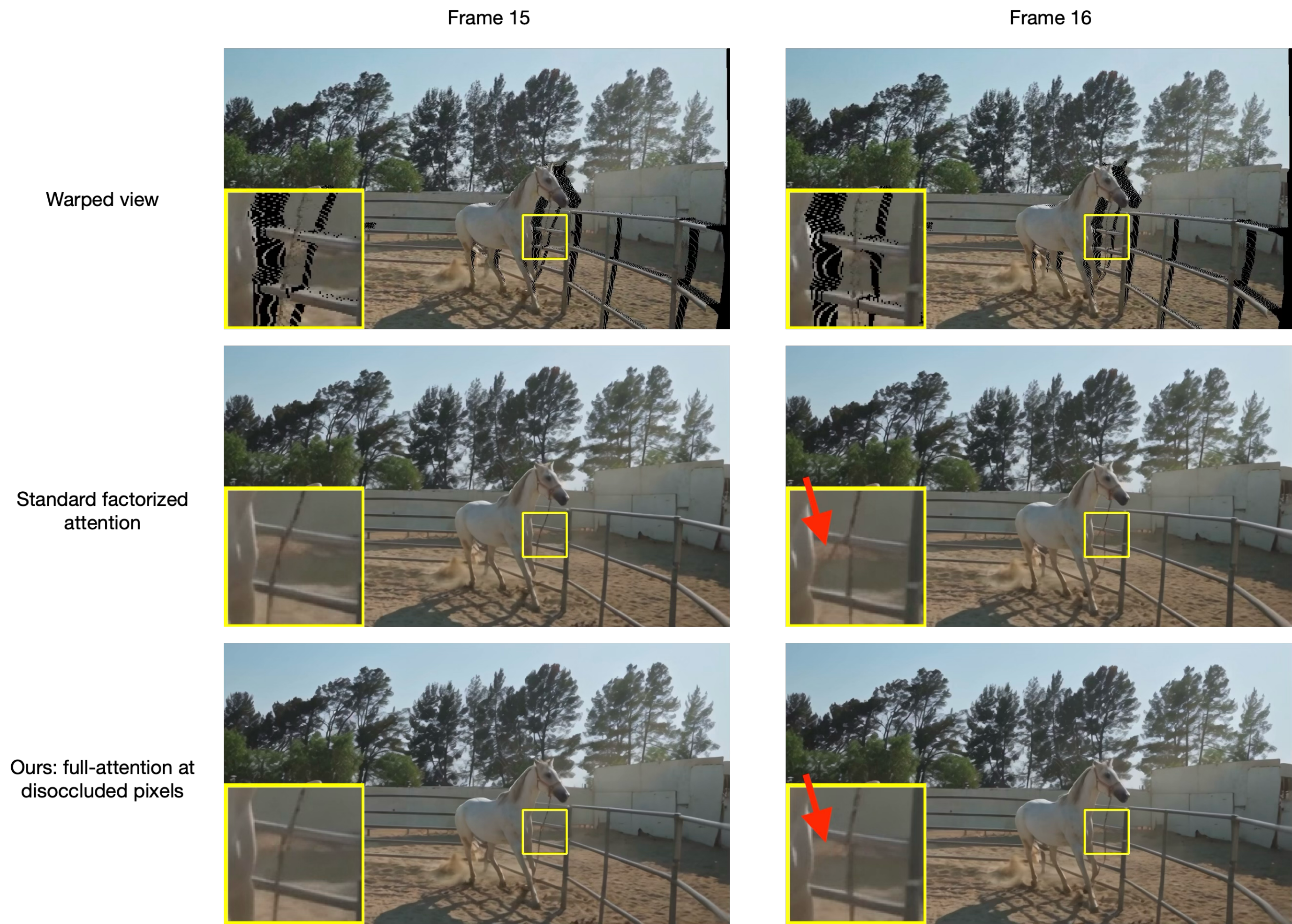}
\vspace{-2mm}
\caption{\small{Standard factorized attention vs full-attention at dis-occluded pixels (ours). Full attention at dis-occluded pixels helps the model better exploit information from other frames. In particular, while standard factorized attention hallucinated the reins at the same spatial location as in the previous frame, creating a visual artifact as if the reins split into two pieces, full attention at dis-occluded pixels prevents this hallucination and correctly inpaints the region. 
\label{fig:alb_full_attention_3}
}}
\end{figure*}

%% file: main.bib
@String(CVPR= {IEEE Conf. Comput. Vis. Pattern Recog.})

@String(ICCV= {Int. Conf. Comput. Vis.})

@String(ECCV= {Eur. Conf. Comput. Vis.})

@String(TIP  = {IEEE Trans. Image Process.})

@String(ICIP = {IEEE Int. Conf. Image Process.})

@String(IJCAI = {IJCAI})

@String(AAAI = {AAAI})

@String(WACV  = {Proceedings of the IEEE/CVF Winter Conference on Applications of Computer Vision})

@String(CVPR  = {CVPR})

@String(ICCV  = {ICCV})

@String(ECCV  = {ECCV})

@String(NeurIPS  = {NeurIPS})

@String(TIP   = {IEEE TIP})

@String(ICIP  = {ICIP})

@String(WACV  = {WACV})

@article{konrad2013learning,
  title={Learning-based, automatic 2D-to-3D image and video conversion},
  author={Konrad, Janusz and Wang, Meng and Ishwar, Prakash and Wu, Chen and Mukherjee, Debargha},
  journal = TIP,
  volume={22},
  number={9},
  pages={3485--3496},
  year={2013},
  publisher={IEEE}
}

@inproceedings{xie2016deep3d,
  title={Deep3d: Fully automatic 2d-to-3d video conversion with deep convolutional neural networks},
  author={Xie, Junyuan and Girshick, Ross and Farhadi, Ali},
  booktitle=ECCV,
  pages={842--857},
  year={2016},
  organization={Springer}
}

@inproceedings{tucker2020single,
  title={Single-view view synthesis with multiplane images},
  author={Tucker, Richard and Snavely, Noah},
  booktitle=CVPR,
  pages={551--560},
  year={2020}
}

@inproceedings{han2022single,
  title={Single-view view synthesis in the wild with learned adaptive multiplane images},
  author={Han, Yuxuan and Wang, Ruicheng and Yang, Jiaolong},
  booktitle={ACM SIGGRAPH 2022 Conference Proceedings},
  pages={1--8},
  year={2022}
}

@inproceedings{wang2023learning,
author = {Wang, Xiaodong and Wu, Chenfei and Yin, Shengming and Ni, Minheng and Wang, Jianfeng and Li, Linjie and Yang, Zhengyuan and Yang, Fan and Wang, Lijuan and Liu, Zicheng and Fang, Yuejian and Duan, Nan},
title = {Learning 3D photography videos via self-supervised diffusion on single images},
year = {2023},
isbn = {978-1-956792-03-4},
url = {https://doi.org/10.24963/ijcai.2023/167},
doi = {10.24963/ijcai.2023/167},
booktitle = {Proceedings of the Thirty-Second International Joint Conference on Artificial Intelligence},
articleno = {167},
numpages = {9},
location = {Macao, P.R.China},
series = {IJCAI '23}
}

@article{shi2024immersepro,
  title={ImmersePro: End-to-End Stereo Video Synthesis Via Implicit Disparity Learning},
  author={Shi, Jian and Li, Zhenyu and Wonka, Peter},
  journal={arXiv preprint arXiv:2410.00262},
  year={2024}
}

@article{Lee2017Automatic2C,
  title={Automatic 2D-to-3D conversion using multi-scale deep neural network},
  author={Jiyoung Lee and Hyungjoo Jung and Youngjung Kim and Kwanghoon Sohn},
  journal=ICIP,
  year={2017},
  url={https://api.semanticscholar.org/CorpusID:3460462}
}

@article{shi2024stereocrafter,
  title={StereoCrafter-Zero: Zero-Shot Stereo Video Generation with Noisy Restart},
  author={Shi, Jian and Wang, Qian and Li, Zhenyu and Wonka, Peter},
  journal={arXiv preprint arXiv:2411.14295},
  year={2024}
}

@InProceedings{spatialdreamer,
  title={SpatialDreamer: Self-supervised Stereo Video Synthesis from Monocular Input},
  author={Zhen Lv and Yangqi Long and Congzhentao Huang and Cao Li and Chengfei Lv and Hao Ren and Dian Zheng},
  booktitle=CVPR,
  year={2025}
}

@InProceedings{single_view_mpi,
  author = {Tucker, Richard and Snavely, Noah},
  title = {Single-view View Synthesis with Multiplane Images},
  booktitle = CVPR,
  year = {2020}
}

@inproceedings{Karras2024edm2,
  title     = {Analyzing and Improving the Training Dynamics of Diffusion Models},
  author    = {Tero Karras and Miika Aittala and Jaakko Lehtinen and
               Janne Hellsten and Timo Aila and Samuli Laine},
  booktitle = {Proc. CVPR},
  year      = {2024},
}

@article{Kawano2024MaskDiffusionEP,
  title={MaskDiffusion: Exploiting Pre-Trained Diffusion Models for Semantic Segmentation},
  author={Yasufumi Kawano and Yoshimitsu Aoki},
  journal={IEEE Access},
  year={2024},
  volume={12},
  pages={127283-127293},
  url={https://api.semanticscholar.org/CorpusID:268513010}
}

@article{Zhang2023AddingCC,
  title={Adding Conditional Control to Text-to-Image Diffusion Models},
  author={Lvmin Zhang and Anyi Rao and Maneesh Agrawala},
  journal=ICCV,
  year={2023},
  url={https://api.semanticscholar.org/CorpusID:256827727}
}

@article{Ramesh2022HierarchicalTI,
  title={Hierarchical Text-Conditional Image Generation with CLIP Latents},
  author={Aditya Ramesh and Prafulla Dhariwal and Alex Nichol and Casey Chu and Mark Chen},
  journal={ArXiv},
  year={2022},
  volume={abs/2204.06125},
  url={https://api.semanticscholar.org/CorpusID:248097655}
}

@article{Saharia2022PhotorealisticTD,
  title={Photorealistic Text-to-Image Diffusion Models with Deep Language Understanding},
  author={Chitwan Saharia and William Chan and Saurabh Saxena and Lala Li and Jay Whang and Emily L. Denton and Seyed Kamyar Seyed Ghasemipour and Burcu Karagol Ayan and Seyedeh Sara Mahdavi and Raphael Gontijo Lopes and Tim Salimans and Jonathan Ho and David J. Fleet and Mohammad Norouzi},
  journal={ArXiv},
  year={2022},
  volume={abs/2205.11487},
  url={https://api.semanticscholar.org/CorpusID:248986576}
}

@inproceedings{fu2024geowizard,
  title={GeoWizard: Unleashing the Diffusion Priors for 3D Geometry Estimation from a Single Image},
  author={Fu, Xiao and Yin, Wei and Hu, Mu and Wang, Kaixuan and Ma, Yuexin and Tan, Ping and Shen, Shaojie and Lin, Dahua and Long, Xiaoxiao},
  booktitle={ECCV},
  year={2024}
}

@article{gao2024cat4d,
    title={CAT4D: Create Anything in 4D with Multi-View Video Diffusion Models},
    author={Rundi Wu and Ruiqi Gao and Ben Poole and Alex Trevithick and Changxi Zheng and Jonathan T. Barron and Aleksander Holynski
    },
    journal={arxiv/2411.18613},
    year={2024}
}

@article{gao2024cat3d,
    title={CAT3D: Create Anything in 3D with Multi-View Diffusion Models},
    author={Ruiqi Gao* and Aleksander Holynski* and Philipp Henzler and Arthur Brussee and Ricardo Martin-Brualla and Pratul P. Srinivasan and Jonathan T. Barron and Ben Poole*
    },
    journal=NeurIPS,
    year={2024}
}

@inproceedings{mildenhall2020nerf,
  title={NeRF: Representing Scenes as Neural Radiance Fields for View Synthesis},
  author={Ben Mildenhall and Pratul P. Srinivasan and Matthew Tancik and Jonathan T. Barron and Ravi Ramamoorthi and Ren Ng},
  year={2020},
  booktitle={ECCV},
}

@misc{kalischek2025cubediffrepurposingdiffusionbasedimage,
      title={CubeDiff: Repurposing Diffusion-Based Image Models for Panorama Generation}, 
      author={Nikolai Kalischek and Michael Oechsle and Fabian Manhardt and Philipp Henzler and Konrad Schindler and Federico Tombari},
      year={2025},
      eprint={2501.17162},
      archivePrefix={arXiv},
      primaryClass={cs.CV},
      url={https://arxiv.org/abs/2501.17162}, 
}

@InProceedings{Muller_2024_CVPR,
                author    = {M\"uller, Norman and Schwarz, Katja and R\"ossle, Barbara and Porzi, Lorenzo and Bul\`o, Samuel Rota and Nie{\ss}ner, Matthias and Kontschieder, Peter},
                title     = {MultiDiff: Consistent Novel View Synthesis from a Single Image},
                booktitle = CVPR,
                year      = {2024},
            }

@inproceedings{Shih3DP20,
  author = {Shih, Meng-Li and Su, Shih-Yang and Kopf, Johannes and Huang, Jia-Bin},
  title = {3D Photography using Context-aware Layered Depth Inpainting},
  booktitle = CVPR,
  year = {2020}
}

@article{Jampani2021SLIDESI,
  title={SLIDE: Single Image 3D Photography with Soft Layering and Depth-aware Inpainting},
  author={V. Jampani and Huiwen Chang and Kyle Sargent and Abhishek Kar and Richard Tucker and Michael Krainin and Dominik Philemon Kaeser and William T. Freeman and D. Salesin and Brian Curless and Ce Liu},
  journal=ICCV,
  year={2021},
  url={https://api.semanticscholar.org/CorpusID:237386323}
}

@article{dai2024svg,
  title={SVG: 3D Stereoscopic Video Generation via Denoising Frame Matrix},
  author={Dai, Peng and Tan, Feitong and Xu, Qiangeng and Futschik, David and Du, Ruofei and Fanello, Sean and Qi, Xiaojuan and Zhang, Yinda},
  journal={arXiv preprint arXiv:2407.00367},
  year={2024}
}

@inproceedings{mehl2024stereo,
  title={Stereo Conversion with Disparity-Aware Warping, Compositing and Inpainting},
  author={Mehl, Lukas and Bruhn, Andr{\'e}s and Gross, Markus and Schroers, Christopher},
  booktitle=WACV,
  year={2024}
}

@misc{chen2024videocrafter2,
      title={VideoCrafter2: Overcoming Data Limitations for High-Quality Video Diffusion Models}, 
      author={Haoxin Chen and Yong Zhang and Xiaodong Cun and Menghan Xia and Xintao Wang and Chao Weng and Ying Shan},
      year={2024},
      eprint={2401.09047},
      archivePrefix={arXiv},
      primaryClass={cs.CV}
}

@article{ho2020denoising,
  title={Denoising diffusion probabilistic models},
  author={Ho, Jonathan and Jain, Ajay and Abbeel, Pieter},
  journal=NeurIPS,
  year={2020}
}

@article{song2020denoising,
  title={Denoising diffusion implicit models},
  author={Song, Jiaming and Meng, Chenlin and Ermon, Stefano},
  journal={arXiv preprint arXiv:2010.02502},
  year={2020}
}

@inproceedings{rombach2022high,
  title={High-resolution image synthesis with latent diffusion models},
  author={Rombach, Robin and Blattmann, Andreas and Lorenz, Dominik and Esser, Patrick and Ommer, Bj{\"o}rn},
  booktitle=CVPR,
  year={2022}
}

@inproceedings{ke2024repurposing,
  title={Repurposing diffusion-based image generators for monocular depth estimation},
  author={Ke, Bingxin and Obukhov, Anton and Huang, Shengyu and Metzger, Nando and Daudt, Rodrigo Caye and Schindler, Konrad},
  booktitle=CVPR,
  year={2024}
}

@article{depth_anything_v2,
  title={Depth Anything V2},
  author={Yang, Lihe and Kang, Bingyi and Huang, Zilong and Zhao, Zhen and Xu, Xiaogang and Feng, Jiashi and Zhao, Hengshuang},
  journal={arXiv:2406.09414},
  year={2024}
}

@article{garcia2024fine,
  title={Fine-tuning image-conditional diffusion models is easier than you think},
  author={Garcia, Gonzalo Martin and Zeid, Karim Abou and Schmidt, Christian and de Geus, Daan and Hermans, Alexander and Leibe, Bastian},
  journal={arXiv preprint arXiv:2409.11355},
  year={2024}
}

@inproceedings{shao2024learningtemporallyconsistentvideo,
  title={Learning temporally consistent video depth from video diffusion priors},
  author={Shao, Jiahao and Yang, Yuanbo and Zhou, Hongyu and Zhang, Youmin and Shen, Yujun and Guizilini, Vitor and Wang, Yue and Poggi, Matteo and Liao, Yiyi},
  booktitle=CVPR,
  year={2025}
}

@inproceedings{grauman2022ego4d,
  title={Ego4d: Around the world in 3,000 hours of egocentric video},
  author={Grauman, Kristen and Westbury, Andrew and Byrne, Eugene and Chavis, Zachary and Furnari, Antonino and Girdhar, Rohit and Hamburger, Jackson and Jiang, Hao and Liu, Miao and Liu, Xingyu and others},
  booktitle=CVPR,
  year={2022}
}

@article{jin2024stereo4d,
  title={Stereo4D: Learning How Things Move in 3D from Internet Stereo Videos},
  author={Jin, Linyi and Tucker, Richard and Li, Zhengqi and Fouhey, David and Snavely, Noah and Holynski, Aleksander},
  journal={arXiv preprint arXiv:2412.09621},
  year={2024}
}

@inproceedings{perazzi2016benchmark,
  title={A benchmark dataset and evaluation methodology for video object segmentation},
  author={Perazzi, Federico and Pont-Tuset, Jordi and McWilliams, Brian and Van Gool, Luc and Gross, Markus and Sorkine-Hornung, Alexander},
  booktitle=CVPR,
  year={2016}
}

@inproceedings{sun2021loftr,
  title={LoFTR: Detector-free local feature matching with transformers},
  author={Sun, Jiaming and Shen, Zehong and Wang, Yuang and Bao, Hujun and Zhou, Xiaowei},
  booktitle=CVPR,
  year={2021}
}

@inproceedings{lee2025videoinpainter,
  title={Video Diffusion Models are Strong Video Inpainter},
  author={Minhyeok Lee and Suhwan Cho and Chajin Shin and Jungho Lee and Sunghun Yang and Sangyoun Lee},
  booktitle={AAAI},
  year={2025}
}

@article{fischler1981random,
  title={Random sample consensus: a paradigm for model fitting with applications to image analysis and automated cartography},
  author={FISCHLER AND, MA},
  journal={Commun. ACM},
  volume={24},
  number={6},
  pages={381--395},
  year={1981}
}

@misc{cherel2023infusion,
      title={Infusion: Internal Diffusion for Video Inpainting}, 
      author={Nicolas Cherel and Andrés Almansa and Yann Gousseau and Alasdair Newson},
      year={2023},
      eprint={2311.01090},
      archivePrefix={arXiv},
      primaryClass={cs.CV}
}

@article{zhang2023avid,
  title={AVID: Any-Length Video Inpainting with Diffusion Model},
  author={Zhang, Zhixing and Wu, Bichen and Wang, Xiaoyan and Luo, Yaqiao and Zhang, Luxin and Zhao, Yinan and Vajda, Peter and Metaxas, Dimitris and Yu, Licheng},
  journal={arXiv preprint arXiv:2312.03816},
  year={2023}
}

@article{Liu2023ImageIV,
  title={Image Inpainting via Tractable Steering of Diffusion Models},
  author={Anji Liu and Mathias Niepert and Guy Van den Broeck},
  journal={ArXiv},
  year={2023},
  volume={abs/2401.03349},
  url={https://api.semanticscholar.org/CorpusID:266196146}
}

@article{Corneanu2024LatentPaintII,
  title={LatentPaint: Image Inpainting in Latent Space with Diffusion Models},
  author={Ciprian Adrian Corneanu and Raghudeep Gadde and Aleix M. Mart{\'i}nez},
  journal=WACV,
  year={2024},
  url={https://api.semanticscholar.org/CorpusID:269035757}
}

@article{Lugmayr2022RePaintIU,
  title={RePaint: Inpainting using Denoising Diffusion Probabilistic Models},
  author={Andreas Lugmayr and Martin Danelljan and Andr{\'e}s Romero and Fisher Yu and Radu Timofte and Luc Van Gool},
  journal=CVPR,
  year={2022},
  url={https://api.semanticscholar.org/CorpusID:246240274}
}

@article{blattmann2023stable,
  title={Stable video diffusion: Scaling latent video diffusion models to large datasets},
  author={Blattmann, Andreas and Dockhorn, Tim and Kulal, Sumith and Mendelevitch, Daniel and Kilian, Maciej and Lorenz, Dominik and Levi, Yam and English, Zion and Voleti, Vikram and Letts, Adam and others},
  journal={arXiv preprint arXiv:2311.15127},
  year={2023}
}

@inproceedings{zhou2023propainter,
  title={Propainter: Improving propagation and transformer for video inpainting},
  author={Zhou, Shangchen and Li, Chongyi and Chan, Kelvin CK and Loy, Chen Change},
  booktitle=ICCV,
  pages={10477--10486},
  year={2023}
}

@article{song2024sdxs,
  author    = {Yuda Song, Zehao Sun, Xuanwu Yin},
  title     = {SDXS: Real-Time One-Step Latent Diffusion Models with Image Conditions},
  journal   = {arxiv},
  year      = {2024},
}

@inproceedings{zhang2018perceptual,
  title={The Unreasonable Effectiveness of Deep Features as a Perceptual Metric},
  author={Zhang, Richard and Isola, Phillip and Efros, Alexei A and Shechtman, Eli and Wang, Oliver},
  booktitle={CVPR},
  year={2018}
}

@article{Mao2024OSVOS,
  title={OSV: One Step is Enough for High-Quality Image to Video Generation},
  author={Xiaofeng Mao and Zhengkai Jiang and Fu-Yun Wang and Wenbing Zhu and Jiangning Zhang and Hao Chen and Mingmin Chi and Yabiao Wang},
  journal={ArXiv},
  year={2024},
  volume={abs/2409.11367},
  url={https://api.semanticscholar.org/CorpusID:272693903}
}

@article{Zhang2023HiPAEO,
  title={HiPA: Enabling One-Step Text-to-Image Diffusion Models via High-Frequency-Promoting Adaptation},
  author={Yifan Zhang and Bryan Hooi},
  journal={ArXiv},
  year={2023},
  volume={abs/2311.18158},
  url={https://api.semanticscholar.org/CorpusID:265506445}
}

@article{Ho2022ImagenVH,
  title={Imagen Video: High Definition Video Generation with Diffusion Models},
  author={Jonathan Ho and William Chan and Chitwan Saharia and Jay Whang and Ruiqi Gao and Alexey A. Gritsenko and Diederik P. Kingma and Ben Poole and Mohammad Norouzi and David J. Fleet and Tim Salimans},
  journal={ArXiv},
  year={2022},
  volume={abs/2210.02303},
  url={https://api.semanticscholar.org/CorpusID:252715883}
}

@inproceedings{li2022towards,
  title={Towards an end-to-end framework for flow-guided video inpainting},
  author={Li, Zhen and Lu, Cheng-Ze and Qin, Jianhua and Guo, Chun-Le and Cheng, Ming-Ming},
  booktitle=CVPR,
  year={2022}
}

@inproceedings{liu2023robust,
  title={Robust dynamic radiance fields},
  author={Liu, Yu-Lun and Gao, Chen and Meuleman, Andreas and Tseng, Hung-Yu and Saraf, Ayush and Kim, Changil and Chuang, Yung-Yu and Kopf, Johannes and Huang, Jia-Bin},
  booktitle=CVPR,
  year={2023}
}

@inproceedings{li2023dynibar,
  title={Dynibar: Neural dynamic image-based rendering},
  author={Li, Zhengqi and Wang, Qianqian and Cole, Forrester and Tucker, Richard and Snavely, Noah},
  booktitle=CVPR,
  year={2023}
}

@article{hu2024depthcrafter,
  title={Depthcrafter: Generating consistent long depth sequences for open-world videos},
  author={Hu, Wenbo and Gao, Xiangjun and Li, Xiaoyu and Zhao, Sijie and Cun, Xiaodong and Zhang, Yong and Quan, Long and Shan, Ying},
  journal={arXiv preprint arXiv:2409.02095},
  year={2024}
}

@inproceedings{mipnerf,
  author    = {Jonathan T. Barron and
               Ben Mildenhall and
               Matthew Tancik and
               Peter Hedman and
               Ricardo Martin{-}Brualla and
               Pratul P. Srinivasan},
  title     = {Mip-NeRF: {A} Multiscale Representation for Anti-Aliasing Neural Radiance
               Fields},
  booktitle = ICCV,
  year      = {2021},
  url       = {https://doi.org/10.1109/ICCV48922.2021.00580},
  doi       = {10.1109/ICCV48922.2021.00580},
  bibsource = {dblp computer science bibliography, https://dblp.org}
}

@inproceedings{mipnerf360,
  author    = {Jonathan T. Barron and
               Ben Mildenhall and
               Dor Verbin and
               Pratul P. Srinivasan and
               Peter Hedman},
  title     = {Mip-NeRF 360: Unbounded Anti-Aliased Neural Radiance Fields},
  booktitle = CVPR,
  year      = {2022},
  url       = {https://doi.org/10.1109/CVPR52688.2022.00539},
  doi       = {10.1109/CVPR52688.2022.00539},
  bibsource = {dblp computer science bibliography, https://dblp.org}
}

@article{mueller2022instant,
    author = {Thomas M\"uller and Alex Evans and Christoph Schied and Alexander Keller},
    title = {Instant Neural Graphics Primitives with a Multiresolution Hash Encoding},
    journal = {ACM Trans. Graph.},
    issue_date = {July 2022},
    volume = {41},
    number = {4},
    month = jul,
    year = {2022},
    pages = {102:1--102:15},
    articleno = {102},
    numpages = {15},
    url = {https://doi.org/10.1145/3528223.3530127},
    doi = {10.1145/3528223.3530127},
    publisher = {ACM},
    address = {New York, NY, USA}
}

@inproceedings{densedepth,
    title={Dense Depth Priors for Neural Radiance Fields from Sparse Input Views}, 
    author={Barbara Roessle and Jonathan T. Barron and Ben Mildenhall and Pratul P. Srinivasan and Matthias Nie{\ss}ner},
    booktitle=CVPR,
    year={2022}
}

@article{nerfmm,
  title={Ne{RF}$--$: Neural Radiance Fields Without Known Camera Parameters},
  author={Zirui Wang and Shangzhe Wu and Weidi Xie and Min Chen and Victor Adrian Prisacariu},
  journal={arXiv preprint arXiv:2102.07064},
  year={2021}
}

@article{sinerf,
  author    = {Yitong Xia and
               Hao Tang and
               Radu Timofte and
               Luc Van Gool},
  title     = {SiNeRF: Sinusoidal Neural Radiance Fields for Joint Pose Estimation
               and Scene Reconstruction},
  journal   = {CoRR},
  volume    = {abs/2210.04553},
  year      = {2022},
  url       = {https://doi.org/10.48550/arXiv.2210.04553},
  doi       = {10.48550/arXiv.2210.04553},
  eprinttype = {arXiv},
  eprint    = {2210.04553},
  bibsource = {dblp computer science bibliography, https://dblp.org}
}

@article{GARF,
  author    = {Shin{-}Fang Chng and
               Sameera Ramasinghe and
               Jamie Sherrah and
               Simon Lucey},
  title     = {{GARF:} Gaussian Activated Radiance Fields for High Fidelity Reconstruction
               and Pose Estimation},
  journal   = {CoRR},
  volume    = {abs/2204.05735},
  year      = {2022},
  url       = {https://doi.org/10.48550/arXiv.2204.05735},
  doi       = {10.48550/arXiv.2204.05735},
  eprinttype = {arXiv},
  eprint    = {2204.05735},
  bibsource = {dblp computer science bibliography, https://dblp.org}
}

@article{NeffSPKMCKS21,
  author    = {Thomas Neff and
               Pascal Stadlbauer and
               Mathias Parger and
               Andreas Kurz and
               Joerg H. Mueller and
               Chakravarty R. Alla Chaitanya and
               Anton Kaplanyan and
               Markus Steinberger},
  title     = {DONeRF: Towards Real-Time Rendering of Compact Neural Radiance Fields
               using Depth Oracle Networks},
  journal   = {Comput. Graph. Forum},
  volume    = {40},
  number    = {4},
  pages     = {45--59},
  year      = {2021},
  url       = {https://doi.org/10.1111/cgf.14340},
  doi       = {10.1111/cgf.14340},
  bibsource = {dblp computer science bibliography, https://dblp.org}
}

@inproceedings{HedmanSMBD21,
  author    = {Peter Hedman and
               Pratul P. Srinivasan and
               Ben Mildenhall and
               Jonathan T. Barron and
               Paul E. Debevec},
  title     = {Baking Neural Radiance Fields for Real-Time View Synthesis},
  booktitle = ICCV,
  year      = {2021},
  url       = {https://doi.org/10.1109/ICCV48922.2021.00582},
  doi       = {10.1109/ICCV48922.2021.00582},
  bibsource = {dblp computer science bibliography, https://dblp.org}
}

@inproceedings{Nex,
  author    = {Suttisak Wizadwongsa and
               Pakkapon Phongthawee and
               Jiraphon Yenphraphai and
               Supasorn Suwajanakorn},
  title     = {NeX: Real-Time View Synthesis With Neural Basis Expansion},
  booktitle = CVPR,
  year      = {2021},
  url       = {https://openaccess.thecvf.com/content/CVPR2021/html/Wizadwongsa\_NeX\_Real-Time\_View\_Synthesis\_With\_Neural\_Basis\_Expansion\_CVPR\_2021\_paper.html},
  doi       = {10.1109/CVPR46437.2021.00843},
  bibsource = {dblp computer science bibliography, https://dblp.org}
}

@inproceedings{PlenOctrees,
  author    = {Alex Yu and
               Ruilong Li and
               Matthew Tancik and
               Hao Li and
               Ren Ng and
               Angjoo Kanazawa},
  title     = {PlenOctrees for Real-time Rendering of Neural Radiance Fields},
  booktitle = ICCV,
  year      = {2021},
  url       = {https://doi.org/10.1109/ICCV48922.2021.00570},
  doi       = {10.1109/ICCV48922.2021.00570},
  bibsource = {dblp computer science bibliography, https://dblp.org}
}

@InProceedings{Regnerf,
  author    = {Michael Niemeyer and Jonathan T. Barron and Ben Mildenhall and Mehdi S. M. Sajjadi and Andreas Geiger and Noha Radwan},  
  title     = {RegNeRF: Regularizing Neural Radiance Fields for View Synthesis from Sparse Inputs},
  booktitle = CVPR,
  year      = {2022},
}

@InProceedings{DSNerf,
    author    = {Deng, Kangle and Liu, Andrew and Zhu, Jun-Yan and Ramanan, Deva},
    title     = {Depth-supervised {NeRF}: Fewer Views and Faster Training for Free},
    booktitle = CVPR,
    month     = {June},
    year      = {2022}
}

@InProceedings{dietnerf,
    author    = {Jain, Ajay and Tancik, Matthew and Abbeel, Pieter},
    title     = {Putting NeRF on a Diet: Semantically Consistent Few-Shot View Synthesis},
    booktitle = ICCV,
    year      = {2021},
}

@inproceedings{barf,
  title={BARF: Bundle-Adjusting Neural Radiance Fields},
  author={Lin, Chen-Hsuan and Ma, Wei-Chiu and Torralba, Antonio and Lucey, Simon},
  booktitle=ICCV,
  year={2021}
}

@inproceedings{GNerF,
  author    = {Quan Meng and
               Anpei Chen and
               Haimin Luo and
               Minye Wu and
               Hao Su and
               Lan Xu and
               Xuming He and
               Jingyi Yu},
  title     = {GNeRF: GAN-based Neural Radiance Field without Posed Camera},
  booktitle = ICCV,
  year      = {2021},
  url       = {https://doi.org/10.1109/ICCV48922.2021.00629},
  doi       = {10.1109/ICCV48922.2021.00629},
  bibsource = {dblp computer science bibliography, https://dblp.org}
}

@inproceedings{colmap,
  author    = {Johannes L. Sch{\"{o}}nberger and
               Jan{-}Michael Frahm},
  title     = {Structure-from-Motion Revisited},
  booktitle = {{CVPR} 2016, Las Vegas, NV, USA},
  pages     = {4104--4113},
  year      = {2016}
}

@inproceedings{reconfusion,
  author       = {Rundi Wu and
                  Ben Mildenhall and
                  Philipp Henzler and
                  Keunhong Park and
                  Ruiqi Gao and
                  Daniel Watson and
                  Pratul P. Srinivasan and
                  Dor Verbin and
                  Jonathan T. Barron and
                  Ben Poole and
                  Aleksander Holynski},
  title        = {ReconFusion: 3D Reconstruction with Diffusion Priors},
  booktitle    = CVPR,
  year         = {2024},
  url          = {https://doi.org/10.1109/CVPR52733.2024.02036},
  doi          = {10.1109/CVPR52733.2024.02036},
  bibsource    = {dblp computer science bibliography, https://dblp.org}
}

@inproceedings{Melas-KyriaziL023,
  author       = {Luke Melas{-}Kyriazi and
                  Iro Laina and
                  Christian Rupprecht and
                  Andrea Vedaldi},
  title        = {RealFusion 360{\textdegree} Reconstruction of Any Object from a Single
                  Image},
  booktitle    = CVPR,
  year         = {2023},
  url          = {https://doi.org/10.1109/CVPR52729.2023.00816},
  doi          = {10.1109/CVPR52729.2023.00816},
  bibsource    = {dblp computer science bibliography, https://dblp.org}
}

@inproceedings{LiuXJCTXS23,
  author       = {Minghua Liu and
                  Chao Xu and
                  Haian Jin and
                  Linghao Chen and
                  Mukund Varma T. and
                  Zexiang Xu and
                  Hao Su},
  editor       = {Alice Oh and
                  Tristan Naumann and
                  Amir Globerson and
                  Kate Saenko and
                  Moritz Hardt and
                  Sergey Levine},
  title        = {One-2-3-45: Any Single Image to 3D Mesh in 45 Seconds without Per-Shape
                  Optimization},
  booktitle    = NeurIPS,
  year         = {2023},
  url          = {http://papers.nips.cc/paper\_files/paper/2023/hash/4683beb6bab325650db13afd05d1a14a-Abstract-Conference.html},
  bibsource    = {dblp computer science bibliography, https://dblp.org}
}

@inproceedings{LiuSCZXW00G024,
  author       = {Minghua Liu and
                  Ruoxi Shi and
                  Linghao Chen and
                  Zhuoyang Zhang and
                  Chao Xu and
                  Xinyue Wei and
                  Hansheng Chen and
                  Chong Zeng and
                  Jiayuan Gu and
                  Hao Su},
  title        = {One-2-3-45++: Fast Single Image to 3D Objects with Consistent Multi-View
                  Generation and 3D Diffusion},
  booktitle    = CVPR,
  year         = {2024},
  url          = {https://doi.org/10.1109/CVPR52733.2024.00960},
  doi          = {10.1109/CVPR52733.2024.00960},
  bibsource    = {dblp computer science bibliography, https://dblp.org}
}

@inproceedings{LiuWHTZV23,
  author       = {Ruoshi Liu and
                  Rundi Wu and
                  Basile Van Hoorick and
                  Pavel Tokmakov and
                  Sergey Zakharov and
                  Carl Vondrick},
  title        = {Zero-1-to-3: Zero-shot One Image to 3D Object},
  booktitle    = ICCV,
  year         = {2023},
  url          = {https://doi.org/10.1109/ICCV51070.2023.00853},
  doi          = {10.1109/ICCV51070.2023.00853},
  bibsource    = {dblp computer science bibliography, https://dblp.org}
}

@article{kerbl20233d,
  title={3d gaussian splatting for real-time radiance field rendering.},
  author={Kerbl, Bernhard and Kopanas, Georgios and Leimk{\"u}hler, Thomas and Drettakis, George},
  journal={ACM Trans. Graph.},
  volume={42},
  number={4},
  pages={139--1},
  year={2023}
}

@InProceedings{Turki_2022_CVPR,
    author    = {Turki, Haithem and Ramanan, Deva and Satyanarayanan, Mahadev},
    title     = {Mega-NERF: Scalable Construction of Large-Scale NeRFs for Virtual Fly-Throughs},
    booktitle = CVPR,
    year      = {2022},
}

@InProceedings{sparf,
          title={SPARF: Neural Radiance Fields from Sparse and Noisy Poses},
          author = {Truong, Prune and Rakotosaona, Marie-Julie and Manhardt, Fabian and Tombari, Federico},
          publisher = CVPR,
          year = {2023}
}

@misc{kingma2013auto,
  title={Auto-encoding variational bayes},
  author={Kingma, Diederik P and Welling, Max and others},
  year={2013},
  publisher={Banff, Canada}
}

@article{saharia2022photorealistic,
  title={Photorealistic text-to-image diffusion models with deep language understanding},
  author={Saharia, Chitwan and Chan, William and Saxena, Saurabh and Li, Lala and Whang, Jay and Denton, Emily L and Ghasemipour, Kamyar and Gontijo Lopes, Raphael and Karagol Ayan, Burcu and Salimans, Tim and others},
  journal=NeurIPS,
  year={2022}
}

@inproceedings{zhang2023adding,
  title={Adding conditional control to text-to-image diffusion models},
  author={Zhang, Lvmin and Rao, Anyi and Agrawala, Maneesh},
  booktitle=ICCV,
  year={2023}
}

@article{jing2024match-stereo-videos,
  title={Match-Stereo-Videos: Bidirectional Alignment for Consistent Dynamic Stereo Matching},
  author={Junpeng Jing and Ye Mao and Krystian Mikolajczyk},
  year={2024}
}

@inproceedings{infonerf,
  author    = {Mijeong Kim and
               Seonguk Seo and
               Bohyung Han},
  title     = {InfoNeRF: Ray Entropy Minimization for Few-Shot Neural Volume Rendering},
  booktitle = CVPR,
  year      = {2022},
  url       = {https://doi.org/10.1109/CVPR52688.2022.01257},
  doi       = {10.1109/CVPR52688.2022.01257},
  bibsource = {dblp computer science bibliography, https://dblp.org}
}

@article{Fishchler,
author = {Fischler, Martin A. and Bolles, Robert C.},
title = {Random Sample Consensus: A Paradigm for Model Fitting with Applications to Image Analysis and Automated Cartography},
year = {1981},
issue_date = {June 1981},
publisher = {Association for Computing Machinery},
address = {New York, NY, USA},
volume = {24},
number = {6},
issn = {0001-0782},
url = {https://doi.org/10.1145/358669.358692},
doi = {10.1145/358669.358692},
journal = {Commun. ACM},
month = jun,
pages = {381–395},
numpages = {15}
}

@Inbook{Zhang2021,
author="Zhang, Zhengyou",
editor="Ikeuchi, Katsushi",
title="Eight-Point Algorithm",
bookTitle="Computer Vision: A Reference Guide",
year="2021",
publisher="Springer International Publishing",
address="Cham",
pages="370--371",
isbn="978-3-030-63416-2",
doi="10.1007/978-3-030-63416-2_156",
url="https://doi.org/10.1007/978-3-030-63416-2_156"
}

@inproceedings{KarHM17,
  author       = {Abhishek Kar and
                  Christian H{\"{a}}ne and
                  Jitendra Malik},
  editor       = {Isabelle Guyon and
                  Ulrike von Luxburg and
                  Samy Bengio and
                  Hanna M. Wallach and
                  Rob Fergus and
                  S. V. N. Vishwanathan and
                  Roman Garnett},
  title        = {Learning a Multi-View Stereo Machine},
  booktitle    = NeurIPS,
  year         = {2017},
  url          = {https://proceedings.neurips.cc/paper/2017/hash/9c838d2e45b2ad1094d42f4ef36764f6-Abstract.html},
  bibsource    = {dblp computer science bibliography, https://dblp.org}
}

@inproceedings{YaoLLFQ18,
  author       = {Yao Yao and
                  Zixin Luo and
                  Shiwei Li and
                  Tian Fang and
                  Long Quan},
  editor       = {Vittorio Ferrari and
                  Martial Hebert and
                  Cristian Sminchisescu and
                  Yair Weiss},
  title        = {MVSNet: Depth Inference for Unstructured Multi-view Stereo},
  booktitle    = ECCV,
  year         = {2018},
  url          = {https://doi.org/10.1007/978-3-030-01237-3\_47},
  doi          = {10.1007/978-3-030-01237-3\_47},
  bibsource    = {dblp computer science bibliography, https://dblp.org}
}

@article{salimans2022progressive,
  title={Progressive distillation for fast sampling of diffusion models},
  author={Salimans, Tim and Ho, Jonathan},
  journal={arXiv preprint arXiv:2202.00512},
  year={2022}
}

@inproceedings{ronneberger2015u,
  title={U-net: Convolutional networks for biomedical image segmentation},
  author={Ronneberger, Olaf and Fischer, Philipp and Brox, Thomas},
  booktitle={Medical image computing and computer-assisted intervention--MICCAI 2015: 18th international conference, Munich, Germany, October 5-9, 2015, proceedings, part III 18},
  pages={234--241},
  year={2015},
  organization={Springer}
}

@article{Blattmann2023StableVD,
  title={Stable Video Diffusion: Scaling Latent Video Diffusion Models to Large Datasets},
  author={A. Blattmann and Tim Dockhorn and Sumith Kulal and Daniel Mendelevitch and Maciej Kilian and Dominik Lorenz},
  journal={ArXiv},
  year={2023},
  volume={abs/2311.15127},
  url={https://api.semanticscholar.org/CorpusID:265312551}
}

@article{chen2016training,
  title={Training deep nets with sublinear memory cost},
  author={Chen, Tianqi and Xu, Bing and Zhang, Chiyuan and Guestrin, Carlos},
  journal={arXiv preprint arXiv:1604.06174},
  year={2016}
}
